\documentclass[10pt,journal,compsoc]{IEEEtran}
\usepackage{graphicx}
\usepackage{graphicx}
\usepackage{amsmath}
\usepackage{amssymb}
\usepackage{booktabs}

\usepackage{algorithm}
\usepackage{algorithmic}
\usepackage{subcaption}

\usepackage{bbm}
\usepackage{xspace}
\usepackage{color}
\makeatletter
\DeclareRobustCommand\onedot{\futurelet\@let@token\@onedot}
\def\@onedot{\ifx\@let@token.\else.\null\fi\xspace}

\def\eg{\emph{e.g}\onedot} 
\def\ie{\emph{i.e}\onedot}

\def\etal{\emph{et al}\onedot}
\makeatother

\ifCLASSOPTIONcompsoc
  \usepackage[nocompress]{cite}
\else
  \usepackage{cite}
\fi
\ifCLASSINFOpdf
\else
\fi
\usepackage{url}
\usepackage{xcolor}
\usepackage{colortbl}
\definecolor{myblue}{RGB}{0,0,0}

\begin{document}
%
% paper title
% Titles are generally capitalized except for words such as a, an, and, as,
% at, but, by, for, in, nor, of, on, or, the, to and up, which are usually
% not capitalized unless they are the first or last word of the title.
% Linebreaks \\ can be used within to get better formatting as desired.
% Do not put math or special symbols in the title.
\title{Exploration and Exploitation of Unlabeled Data for Open-Set Semi-Supervised Learning}
%
%
% author names and IEEE memberships
% note positions of commas and nonbreaking spaces ( ~ ) LaTeX will not break
% a structure at a ~ so this keeps an author's name from being broken across
% two lines.
% use \thanks{} to gain access to the first footnote area
% a separate \thanks must be used for each paragraph as LaTeX2e's \thanks
% was not built to handle multiple paragraphs
%
%
%\IEEEcompsocitemizethanks is a special \thanks that produces the bulleted
% lists the Computer Society journals use for "first footnote" author
% affiliations. Use \IEEEcompsocthanksitem which works much like \item
% for each affiliation group. When not in compsoc mode,
% \IEEEcompsocitemizethanks becomes like \thanks and
% \IEEEcompsocthanksitem becomes a line break with idention. This
% facilitates dual compilation, although admittedly the differences in the
% desired content of \author between the different types of papers makes a
% one-size-fits-all approach a daunting prospect. For instance, compsoc 
% journal papers have the author affiliations above the "Manuscript
% received ..."  text while in non-compsoc journals this is reversed. Sigh.

\author{Ganlong~Zhao, 
Guanbin~Li, %~\IEEEmembership{Member,~IEEE,}
Yipeng~Qin, 
Jinjin~Zhang, 
Zhenhua~Chai,
Xiaolin Wei,
Liang~Lin, %~\IEEEmembership{Senior Member,~IEEE,}
Yizhou~Yu%~\IEEEmembership{Fellow,~IEEE,}

\IEEEcompsocitemizethanks{
    \IEEEcompsocthanksitem G. Zhao, G. Li and L. Lin are with the School of Computer Science and Engineering, Sun Yat-sen University, Guangzhou, 510006, China (e-mail: zhaogl@connect.hku.hk; liguanbin@mail.sysu.edu.cn; linliang@ieee.org).
    \IEEEcompsocthanksitem Y. Qin is with the School of Computer Science and Informatics, Cardiff University (e-mail:QinY16@cardiff.ac.uk).
    \IEEEcompsocthanksitem J. Zhang, Z. Chai and X. Wei are the MeituanDianping group. (e-mail: zhangjinjin05@meituan.com; chaizhenhua@meituan.com; weixiaolin02@meituan.com).
    \IEEEcompsocthanksitem Y. Yu is with the Department
of Computer Science, The University of Hong Kong, Hong Kong (e-mail: yizhouy@acm.org).
    \IEEEcompsocthanksitem Corresponding author: Guanbin Li.
	%\IEEEcompsocthanksitem XXX is with the XXXXXXXXXXXXXXXXXXXXXXXXXXX
	}

% \IEEEcompsocitemizethanks{\IEEEcompsocthanksitem M. Shell was with the Department
% of Electrical and Computer Engineering, Georgia Institute of Technology, Atlanta,
% GA, 30332.\protect\\
% % note need leading \protect in front of \\ to get a newline within \thanks as
% % \\ is fragile and will error, could use \hfil\break instead.
% E-mail: see http://www.michaelshell.org/contact.html
% \IEEEcompsocthanksitem J. Doe and J. Doe are with Anonymous University.}% <-this % stops an unwanted space
%\thanks{Manuscript received April 19, 2005; revised August 26, 2015.}
}

% note the % following the last \IEEEmembership and also \thanks - 
% these prevent an unwanted space from occurring between the last author name
% and the end of the author line. i.e., if you had this:
% 
% \author{....lastname \thanks{...} \thanks{...} }
%                     ^------------^------------^----Do not want these spaces!
%
% a space would be appended to the last name and could cause every name on that
% line to be shifted left slightly. This is one of those "LaTeX things". For
% instance, "\textbf{A} \textbf{B}" will typeset as "A B" not "AB". To get
% "AB" then you have to do: "\textbf{A}\textbf{B}"
% \thanks is no different in this regard, so shield the last } of each \thanks
% that ends a line with a % and do not let a space in before the next \thanks.
% Spaces after \IEEEmembership other than the last one are OK (and needed) as
% you are supposed to have spaces between the names. For what it is worth,
% this is a minor point as most people would not even notice if the said evil
% space somehow managed to creep in.

% The paper headers
\markboth{Journal of \LaTeX\ Class Files,~Vol.~14, No.~8, August~2015}%
{Shell \MakeLowercase{\textit{et al.}}: Bare Demo of IEEEtran.cls for Computer Society Journals}
% The only time the second header will appear is for the odd numbered pages
% after the title page when using the twoside option.
% 
% *** Note that you probably will NOT want to include the author's ***
% *** name in the headers of peer review papers.                   ***
% You can use \ifCLASSOPTIONpeerreview for conditional compilation here if
% you desire.

% The publisher's ID mark at the bottom of the page is less important with
% Computer Society journal papers as those publications place the marks
% outside of the main text columns and, therefore, unlike regular IEEE
% journals, the available text space is not reduced by their presence.
% If you want to put a publisher's ID mark on the page you can do it like
% this:
%\IEEEpubid{0000--0000/00\$00.00~\copyright~2015 IEEE}
% or like this to get the Computer Society new two part style.
%\IEEEpubid{\makebox[\columnwidth]{\hfill 0000--0000/00/\$00.00~\copyright~2015 IEEE}%
%\hspace{\columnsep}\makebox[\columnwidth]{Published by the IEEE Computer Society\hfill}}
% Remember, if you use this you must call \IEEEpubidadjcol in the second
% column for its text to clear the IEEEpubid mark (Computer Society jorunal
% papers don't need this extra clearance.)

% use for special paper notices
%\IEEEspecialpapernotice{(Invited Paper)}

% for Computer Society papers, we must declare the abstract and index terms
% PRIOR to the title within the \IEEEtitleabstractindextext IEEEtran
% command as these need to go into the title area created by \maketitle.
% As a general rule, do not put math, special symbols or citations
% in the abstract or keywords.
\IEEEtitleabstractindextext{%
\begin{abstract}
    % In this paper, we address a complex but practical scenario in semi-supervised learning~(SSL) named open-set SSL, where out-of-distribution (OOD) samples are contained in unlabeled data.
    In this paper, we address a complex but practical scenario in semi-supervised learning~(SSL) named open-set SSL, where unlabeled data contain both in-distribution (ID) and out-of-distribution (OOD) samples. 
    % Instead of identifying and completely filtering out OOD samples during training as previous methods do, we find that properly exploiting the presence of OOD data while avoiding its interference with in-distribution (ID) samples can enhance feature learning and benefit SSL.
    % Unlike previous methods that only consider ID samples to be useful and aim to filter out  OOD samples completely during training, we argue that the exploration and exploitation of ID together with OOD samples can benefit SSL.
    Unlike previous methods that only consider ID samples to be useful and aim to filter out OOD ones completely during training, we argue that the exploration and exploitation of \textit{\textbf{both}} ID and OOD samples can benefit SSL.
    % To achieve this goal, we first propose a prototype-based clustering algorithm which \textit{\textbf{explores}} the inherent similarity and difference among samples at the feature level to effectively distinguish between ID and OOD samples.
    To support our claim, i) we propose a prototype-based clustering and identification algorithm that \textit{\textbf{explores}} the inherent similarity and difference among samples at feature level and effectively cluster them around several predefined ID and OOD prototypes, thereby enhancing feature learning and facilitating ID/OOD identification;
    ii) we propose an importance-based sampling method that \textit{\textbf{exploits}} the difference in importance of each ID and OOD sample to SSL, thereby reducing the sampling bias and improving the training.
    % Moreover, we propose a sample refinement strategy which \textit{\textbf{exploits}} the clustering results to identify ID and OOD samples and distill ID samples from unlabeled data to facilitate the training.
    Our proposed method achieves state-of-the-art in several challenging benchmarks, and improves upon existing SSL methods even when ID samples are totally absent in unlabeled data.
\end{abstract}

% Note that keywords are not normally used for peerreview papers.
}

% make the title area
\maketitle

% To allow for easy dual compilation without having to reenter the
% abstract/keywords data, the \IEEEtitleabstractindextext text will
% not be used in maketitle, but will appear (i.e., to be "transported")
% here as \IEEEdisplaynontitleabstractindextext when the compsoc 
% or transmag modes are not selected <OR> if conference mode is selected 
% - because all conference papers position the abstract like regular
% papers do.
\IEEEdisplaynontitleabstractindextext
% \IEEEdisplaynontitleabstractindextext has no effect when using
% compsoc or transmag under a non-conference mode.

% For peer review papers, you can put extra information on the cover
% page as needed:
% \ifCLASSOPTIONpeerreview
% \begin{center} \bfseries EDICS Category: 3-BBND \end{center}
% \fi
%
% For peerreview papers, this IEEEtran command inserts a page break and
% creates the second title. It will be ignored for other modes.
\IEEEpeerreviewmaketitle

\IEEEraisesectionheading{\section{Introduction}\label{sec:introduction}}

Semi-supervised learning (SSL) is a promising machine learning approach that exploits unlabeled data to mitigate the costly data labeling process. 
Given a small set of labeled data and a large set of unlabeled data, SSL aims to train a classifier that surpasses its supervised variant trained only on the labeled dataset. 
Classic SSL techniques include consistency regularization~\cite{sajjadi2016regularization, laine2016temporal}, pseudo labeling ({\it a.k.a.} self-training)~\cite{lee2013pseudo,pham2021meta} and entropy minimization~\cite{grandvalet2004semi}.
Recently, FixMatch~\cite{sohn2020fixmatch} achieved state-of-the-art performance by simply combining consistency regularization with pseudo labeling.
Although being effective, traditional SSL methods implicitly assumed that the unlabeled data share the same label space with the labeled data during training, which limits their application in the open-set real-world scenarios.

\begin{figure}[t]
  \begin{center}
      \includegraphics[width=\linewidth]{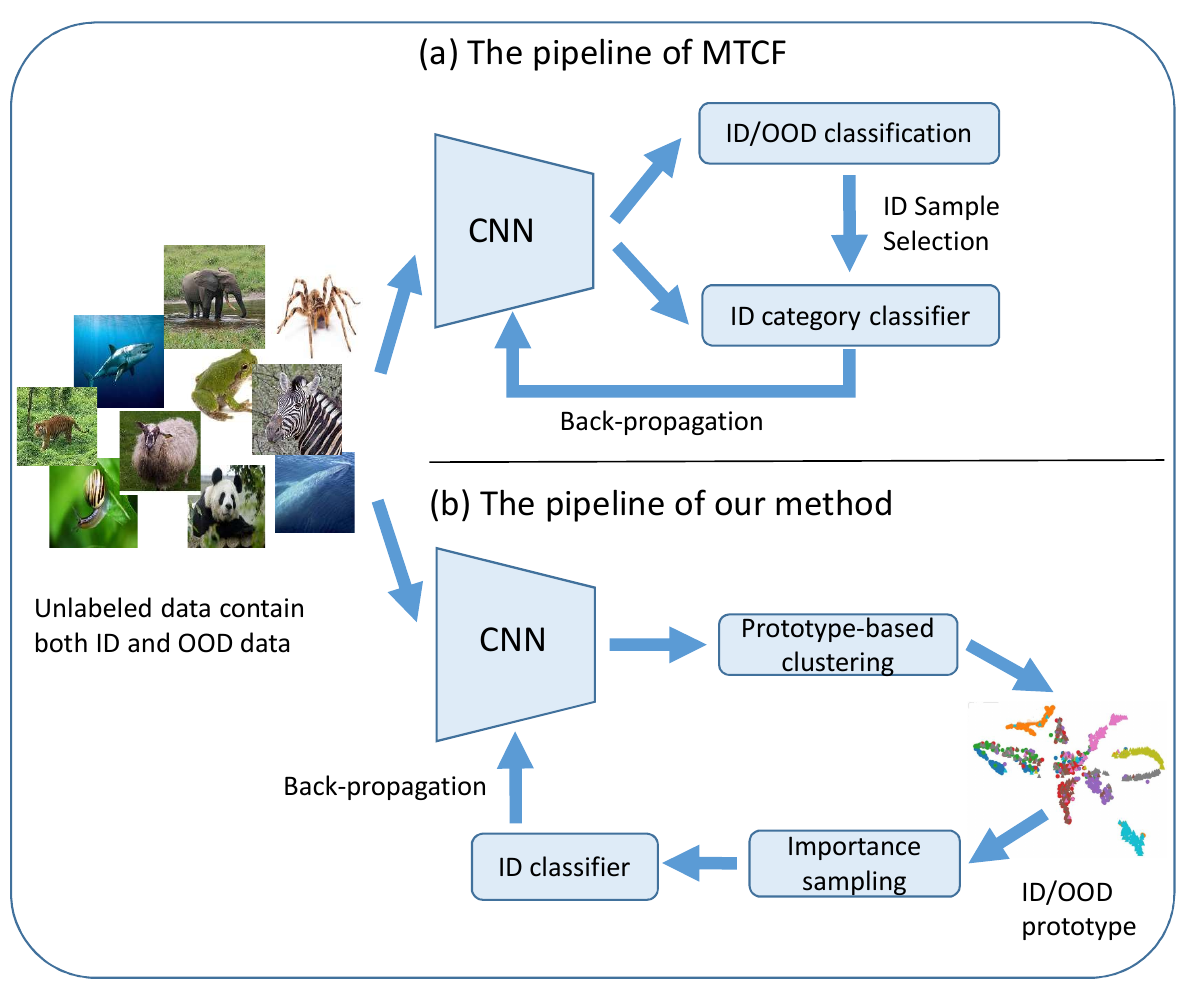}
  \end{center}
  \caption{The comparison between our method and MTCF\cite{yu2020multi}. MTCF sets an independent OOD detection branch to the backbone, and performs ID/OOD classification on the features. OOD samples are excluded from semi-supervised learning. Different from MTCF, our method first clusters both ID and OOD features to prototypes. Prototypes are assigned ID/OOD labels and unlabeled samples are sampled by their importance and prototypes. Our method exploits both ID and OOD data for the backbone training.}
  \label{motivation_figure}
\end{figure}

\begin{figure*}[t]
  \begin{center}
      \includegraphics[width=\linewidth]{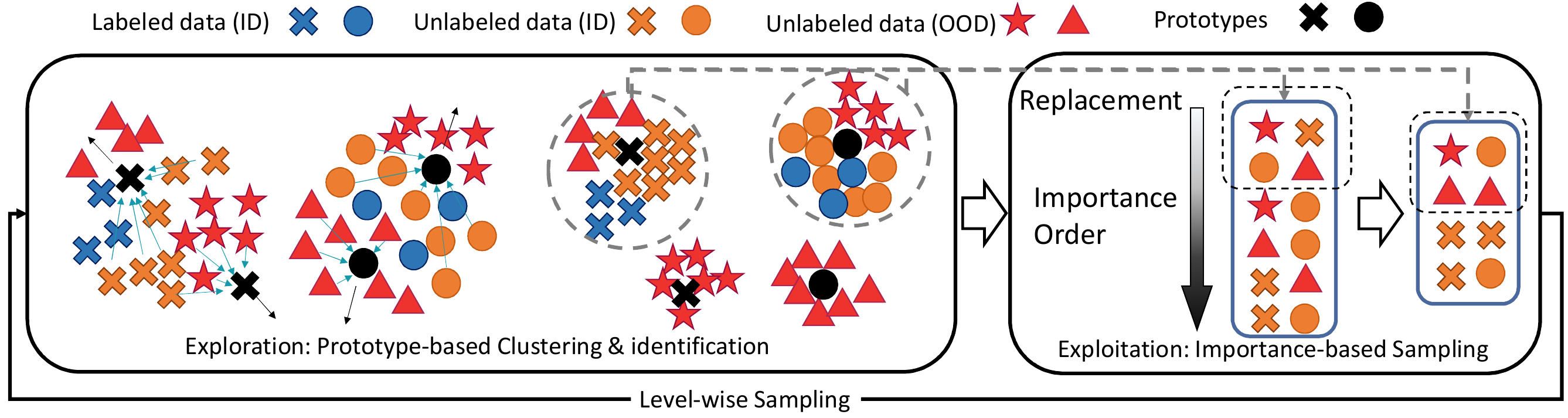}
  \end{center}
  \caption{Exploration and exploitation of the unlabeled data including in-distribution (ID) and out-of-distribution (OOD) samples. 
  Exploration: we cluster ID and OOD samples in each mini-batch and identify them accordingly based on an exploration of the inherent similarity and difference of their features (represented by prototypes).
  Exploitation: we exploit the different importance of each ID and OOD sample in SSL to reduce the sampling bias during training.}
  \label{motivation_figure}
\end{figure*}

Open-set SSL extends SSL to open-set datasets where the unlabeled data contain both in-distribution (ID) and out-of-distribution (OOD) samples. 
Specifically, ID samples share the same label space with labeled data while OOD samples may be out of that label space.
Yu et al.\cite{yu2020multi} pioneered this direction and proposed to eliminate the negative effects of OOD samples using an OOD detector~\cite{liang2018enhancing}.
Thanks to the OOD detector, they identified high-confidence ID samples and gradually incorporated them into the training of a MixMatch~\cite{berthelot2019mixmatch} model with their multi-task curriculum framework.

Although MTCF \cite{yu2020multi} is effective, we argue that it has the following two shortcomings.
First, it overlooks the role of OOD samples in feature learning.
In their method, OOD samples are excluded from SSL whereas we argue that {\it if being properly used, OOD samples can benefit feature learning and thus SSL}, especially when there are few ID samples in the unlabelled dataset.
Second, their method depends on the performance of its OOD detector and thus performs poorly on high-variance datasets where the ambiguity between ID and OOD samples makes it prone to misclassification. 
As pointed out by previous studies\cite{winkens2020contrastive}, near-OOD tasks where OOD samples are close to ID ones can greatly lower the performance of OOD detection method.
Simply filtering out all OOD samples can be difficult and thus degrades the performance of semi-supervised training when OOD samples dominate the unlabeled dataset.
In addition, their evaluation is based on synthetic OOD samples (\eg Gaussian noise, Uniform noise) and images of completely irrelevant topics, which may not generalize to real-world scenarios where OOD samples can be ``close'' to ID ones.

In previous semi-supervised studies, pseudo labeling is an important technique that can utilize the unlabeled data and thus improve the performance of semi-supervised methods. Pseudo labeling encourages the model to output high-confidence prediction for unlabeled samples and thus construct a better feature extractor\cite{sohn2020fixmatch}. However, if unlabeled data contains both ID and OOD images, pseudo-labeling-based methods will force ID and OOD samples with the same label prediction to get closer, which degrades the performance of the feature extractor and the accuracy of ID/OOD classification. Therefore, our method aims to construct and preserve the inner structure of both ID and OOD features to train a better feature extractor and to facilitate the ID/OOD classification.

In this paper, we address the aforementioned shortcomings of open-set SSL by exploring and exploiting the unlabeled data including both ID and OOD samples (Fig.~\ref{motivation_figure}).
Specifically, we first propose a prototype-based clustering and identification algorithm that clarifies the ambiguity between ID and OOD samples by {\it exploring} the inherent similarity and difference among their features, and thus better identifies the unlabeled samples.
Then, we propose a novel importance sampling method that reduces the sampling bias by {\it exploiting} the difference in importance of each ID and OOD sample to SSL, thereby improving training. 
We implement this method with our newly proposed cascading pooling strategy, which increases the density of ID samples in mini-batches and further stabilizes training.
Empirically, we verify the effectiveness of our method on three standard benchmark datasets (CIFAR-100~\cite{krizhevsky2009learning}, SVHN~\cite{netzer2011reading} and TinyImageNet~\cite{deng2009imagenet}) and a new dataset, DomainNet-Real~\cite{peng2019moment}, which is more challenging and realistic.
In summary, our contributions include:
\begin{itemize}
    \item We demonstrate that the performance of open-set semi-supervised learning (SSL) can be improved by utilizing out-of-distribution (OOD) samples.
    \item We design a novel prototype-based clustering and identification algorithm and demonstrate its effectiveness in feature learning.
    \item We propose a new importance-based sampling method that reduces sampling bias and improves training.
    \item We introduce a new benchmark that is more challenging and closer to real-world scenarios. \textcolor{myblue}{Extensive experimental results on three standard benchmark datasets as well as our introduced benchmark demonstrate the superiority of our proposed method.}
\end{itemize}

\begin{figure*}[t]
   \begin{center}
      \includegraphics[width=\linewidth]{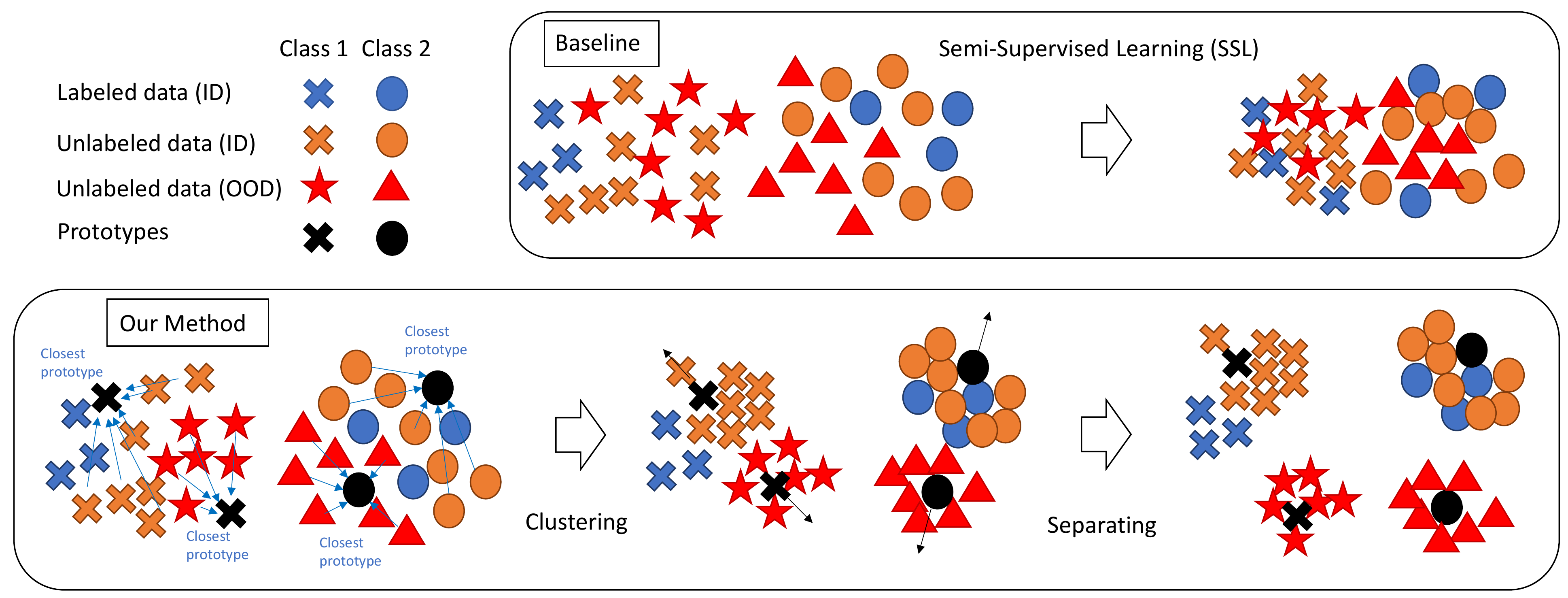}
   \end{center}
    \caption{Illustration of our prototype-based clustering method. 
    Top row: Semi-supervised learning (SSL) methods might be confused by the OOD samples and incorrectly assign in-distribution (ID) pseudo labels to them, which degrades their performance. 
    Bottom row: With our clustering method, ID and OOD samples are pushed away from each other towards a set of pre-defined prototypes (black marks), which clarifies the ambiguity between ID and OOD samples and facilitates ID/OOD identification. The positions of the prototypes are dynamically updated during training.}
   \label{clustering_figure}
\end{figure*}

\section{Related Works}

\noindent {\bf Semi-Supervised Learning (SSL)} addresses the scarcity of labeled data by leveraging the relationship between a small amount of labeled data and a large amount of unlabeled data. 
In general, two common SSL techniques that are widely applied to semi-supervised learning are consistency regularization and pseudo labeling ({\it a.k.a.}~self-training). 
Consistency regularization (CR)~\cite{laine2016temporal,bachman2014learning, sajjadi2016regularization, zhai2019s4l} assumes that the classification results should only rely on the semantics of input images, and penalizes the change of model outputs against the perturbation or augmentation of input images. 
Some CR methods employ adversarial perturbation or dropout~\cite{park2018adversarial, wager2013dropout} on the input images while data augmentation~\cite{berthelot2019remixmatch, sajjadi2016regularization} is widely recognized to be more effective. 
From another perspective, pseudo labeling~\cite{lee2013pseudo,pham2021meta} assigns pseudo labels to unlabeled data according to the model's prediction confidence and steers its own training with those pseudo labels. FixMatch\cite{sohn2020fixmatch} combines the ideas of pseudo labeling and consistency regularization, and achieves state-of-the-art performance on several benchmarks for semi-supervised learning. FixMatch utilizes two different augmentations of the input image, strong augmentation, and weak augmentation, and trains the model with the strong-augmented images and the pseudo labels generated by corresponding weak-augmented images.
Similar to pseudo labeling, Entropy minimization~\cite{grandvalet2004semi} encourages the model to output low-entropy ({\it i.e.} high confidence) prediction for unlabeled samples.
Besides, there have been other techniques for semi-supervised learning. Temporal ensemble\cite{laine2016temporal} forms a consensus prediction for the unlabeled data using the outputs of the network-in-training on different epochs. Mean teacher\cite{tarvainen2017mean} averages model weights instead of label predictions to avoid the problem that temporal ensemble becomes unwieldy when learning from large datasets.
\textcolor{myblue}{FlexMatch\cite{zhang2021flexmatch} proposes a curriculum learning approach for semi-supervised learning to leverage unlabeled data according to the model's learning status. Some self-supervised methods\cite{li2020prototypical} also employ prototype-based methods for semi-supervised learning, however, their clustering strategies are purely unsupervised and not applicable to OOD detection during training.}

\vspace*{2mm}
\noindent {\bf Out-Of-Distribution (OOD) Identification}~\textcolor{myblue}{\cite{liang2018enhancing, devries2018learning,ming2022delving,du2022siren,yangopenood}} aims to identify the OOD samples in a given dataset which consists of both In-Distribution classes and Out-Of-Distribution samples.
For image classification, conventional methods like density estimation or nearest neighbor~\cite{chow1970optimum,vincent2003manifold,ghoting2008fast} are not applicable due to the high dimensionality of image feature space.
Addressing this issue, DNN-based OOD detectors~\cite{liang2018enhancing, hendrycks2016baseline} have been proposed.
Based on the observation that ID samples tend to have higher softmax scores, Hendrycks et al.~\cite{hendrycks2016baseline} propose a baseline method for OOD detection without retraining networks. 
Liang et al.~\cite{liang2018enhancing} improve such a baseline by introducing temperature scaling in the softmax function to increase the softmax score gap between ID and OOD samples. 
The difficulty of the OOD detection depends on how semantically close to the inlier classes, \ie, ID classes are to the outliers, \ie, OOD samples. Winkens \etal\cite{winkens2020contrastive} distinguish the difficulty difference between near-OOD tasks and far-OOD tasks by the difference of state-of-the-art performance for area under the receiver operating characteristic curve (AUROC).
\textcolor{myblue}{Some methods\cite{lee2018simple,liu2020energy,hsu2020generalized,sun2021react} tackle the OOD detection problem by class conditional Gaussian distributions, energy function or rectified activations. However, most of them detect the OOD samples post hoc, which is not suitable for open-set semi-supervised learning.}

\vspace*{2mm}
\noindent {\bf Open-Set Semi-Supervised Learning} aims to develop robust SSL algorithms which work on ``dirty'' unlabeled data that contain OOD samples. 
Oliver et al.~\cite{oliver2018realistic} first pointed out that the performance of SSL techniques can degrade drastically when the unlabeled data contain a different distribution of classes.
This inspires MTCF~\cite{yu2020multi} which incorporates an OOD detection branch to MixMatch~\cite{berthelot2019mixmatch} and works by gradually adding high-confidence ID samples to semi-supervised training.
However, it ignores the contribution of consistency regularization to SSL, which is independent to OOD detection.
From this perspective, OOD samples are harmless and can even be beneficial.
Thus, excluding them from training may not be the optimal solution and can be impractical for big datasets containing a large proportion of OOD samples.
To this end, we propose to utilize OOD samples instead of filtering them out during training.
As a concurrent work, Luo et al.~\cite{luo2021consistency} viewed the categorical difference between OOD and ID samples as a distributional difference and attempted to reduce the distribution divergence using style transfer. They also explored the OOD samples during training via unsupervised data augmentation~\cite{xie2019unsupervised}.
UASD~\cite{chen2020semi} tackled a problem called {\it Class Distribution MisMatch} where some classes in the labeled data are absent in the unlabeled data, and vice versa. Although looks similar, this problem is different from ours. Huang \etal\cite{huang2021trash} propose a cross-modal matching strategy to detect OOD samples and train the network to match samples to an assigned one-hot class label.
% In addition, instead of utilizing OOD samples, they remove them from the training using the SSL outputs.

\section{Preliminary}

\textcolor{myblue}{Given a small labeled dataset $\mathcal{X} = \{(x_i, y_i)\}_{i=1}^{N_l}$ and a large unlabeled dataset $\hat{\mathcal{X}}=\{\hat{x}_i\}_{i=1}^N$ where $N \gg N_l$ and $y_i \in (1, ... S)$, semi-supervised learning for classification aims to learn a model that performs best by utilizing both $\mathcal{X}$ and $\hat{\mathcal{X}}$. Different from traditional semi-supervised learning, open-set semi-supervised learning aims to utilize an unlabeled dataset $\hat{\mathcal{X}}$ containing out-of-distribution samples whose ground truth labels are not in $(1, ... S)$. Our method aims to train a model to achieve higher accuracy on the test set which contains only in-distribution data.}

\section{Method}

\begin{figure*}[t]
   \begin{center}
      \includegraphics[width=\linewidth]{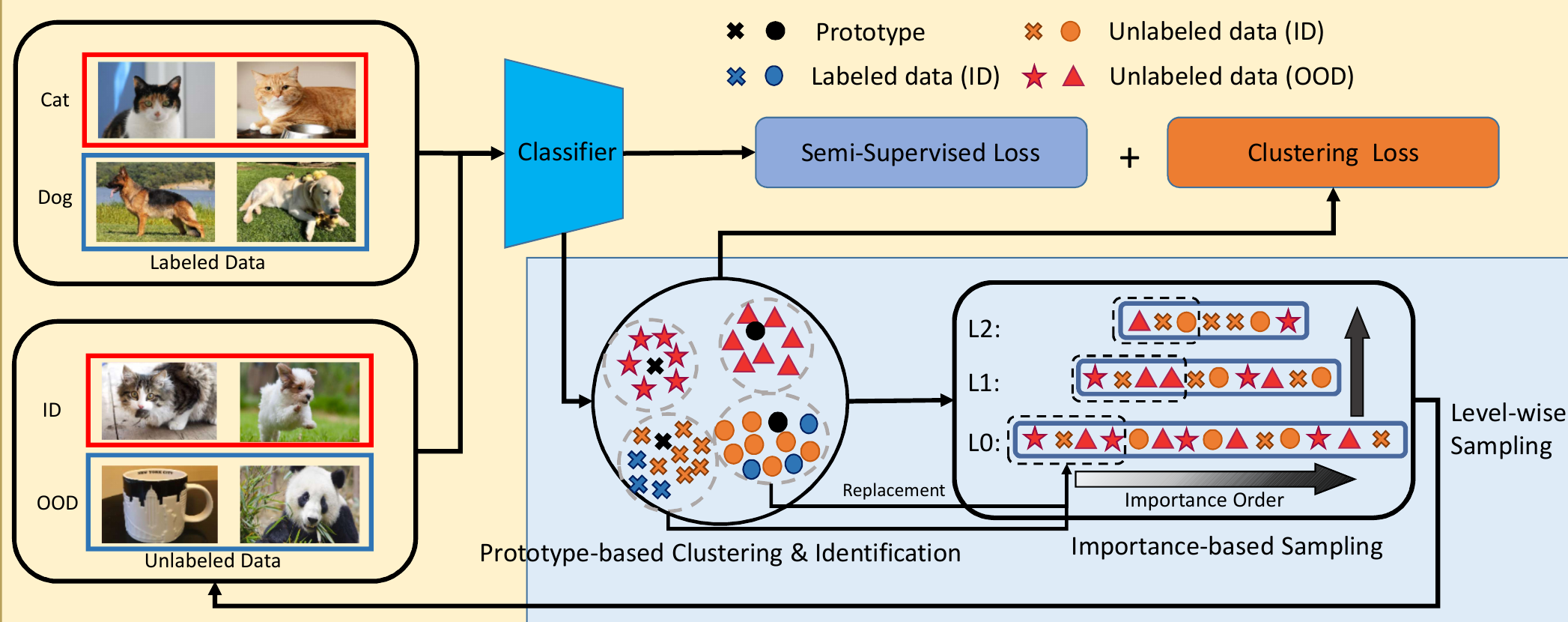}
   \end{center}
    \caption{Overview of our method. We propose two techniques, a prototype-based clustering and identification algorithm,  and an importance-based sampling method to improve the performance of open-set semi-supervised learning (SSL).
    Our clustering and identification algorithm clusters samples at the feature level and thus facilitates feature learning by increasing the distances between ID and OOD samples.
    Our importance-based sampling method facilitates SSL by reducing the sampling bias during training.}
   \label{overview}
\end{figure*}

Our method has two components: i) a prototype-based clustering and identification algorithm that learns better representations for the identification of In-Distribution (ID) and Out-Of-Distribution (OOD) samples by clustering them in an unsupervised way; ii) an importance sampling method that samples unlabeled data according to their importance to SSL, thereby reducing the sampling bias and improving the training (Fig.~\ref{overview}). 
Specifically, our clustering and identification algorithm helps pseudo-labeling by pushing ambiguous ID and OOD samples away from each other (towards different prototypes) in the feature space.
Note that as an unsupervised representation learning method, our clustering process benefits a lot from the OOD data that ``augment'' the dataset.
The resulting clusters can be binarily identified as ID and OOD ones according to labeled data.
Based on the identification, we design a novel importance sampling method that assigns importance scores to unlabeled data and samples them accordingly.
This addresses the problem of random sampling where early-identified ID samples are over-sampled while later ones are under-sampled.
Furthermore, we devise a cascading pooling strategy to improve the density of ID samples in mini-batch training, which further stabilizes the training. 
% At each level, ID samples are dynamically updated for the purpose of hard sample mining.
% Thanks to the relatively large proportions of ID samples with refinement strategy, the density of ID samples in a mini-batch is increased. Therefore, the gradients during training become more stable and the model can focus on ID samples classification. 
The overview of our method is shown in Algorithm~\ref{code:recentEnd}. 
% The SSL part is not included for clarity.

\begin{algorithm}[h]
% \caption{Overview of our method. The SSL part is not included for clarity.}
\caption{Overview of our method.}
\begin{algorithmic}[1]

\REQUIRE %Input
initialized prototypes
\ENSURE clustering loss $L_c$, \textcolor{myblue}{semi-supervised loss $L_{SSL}$}

\FOR{number of training iterations}
% \STATE Sample a minibatch of unlabeled samples from our pyramid of sample pools (Sec.~\ref{sec:sample_refinement})
\STATE \textcolor{myblue}{Sample a minibatch of labeled samples}
\STATE Sample a minibatch of unlabeled samples from our pyramid of sample pools
\STATE \textcolor{myblue}{Compute semi-supervised learning loss $L_{SSL}$ for labeled and unlabeled minibatches}
\FOR{each sample}
\FOR{each $s \in \{1,2,...,S\}$}
% \STATE Compute $L_c$ (Eq.~\ref{eq:L_cluster_bath}) using our prototype-based clustering algorithm (Sec.~\ref{sec:clustering})
\STATE Compute $L_c$ (Eq.~\ref{eq:L_cluster_bath}) using our prototype-based clustering algorithm
% \STATE Update the prototypes using Eq.~\ref{equation_prototype_update} (Sec.~\ref{sec:clustering})
\STATE Update the prototypes using Eq.~\ref{equation_prototype_update}
% \STATE Update the sample pools accordingly (Sec.~\ref{sec:sample_refinement})
\STATE Update the sample pools accordingly 
\ENDFOR
\ENDFOR
\ENDFOR
% \STATE Update the model by minimizing $L_{SSL}$ and $L_c$
\end{algorithmic}
\label{code:recentEnd}
\end{algorithm}

\subsection{Prototype-based Clustering and Identification}
\label{sec:clustering}

As Fig.~\ref{clustering_figure} (top row) shows, ID and OOD samples are mixed in unlabeled data. 
Thus, a key defect of open-set pseudo labeling is that OOD samples can easily be misclassified as ID samples, thereby confusing the feature extractor.
Addressing this issue, we propose a prototype-based clustering algorithm to clarify the ambiguity between ID and OOD samples (Fig.~\ref{clustering_figure}, bottom row).
Let $F$ be an SSL classifier, $\hat{x}_i$ be a sample in the unlabeled dataset $\hat{\mathcal{X}}=\{\hat{x}_i\}_{i=1}^N$, \textcolor{myblue}{$\hat{s}$} be the pseudo label of $\hat{x}_i$ assigned by $F$, $F(\hat{x}_i)$ be the output probabilities of $F$ with input $\hat{x}_i$, and $f(\hat{x}_i)$ be the \textcolor{myblue}{normalized} feature extracted by $f$ (a subnetwork of $F$, {\it a.k.a.} a feature extractor), our clustering algorithm is detailed as follows:

\vspace*{2mm}
\noindent {\bf Prototype Initialization.}
This step aims to set up $K$ initial prototypes $p^s_j, j \in \{1,2,...,K\}$ for each class $s \in \{1,2,...,S\}$.
First, we pretrain $F$ until each class $s$ contains at least $M$ unlabelled samples, $M\ll N/S$ \textcolor{myblue}{where $N$ is the unlabeled set size. }
These samples are assigned pseudo labels \textcolor{myblue}{$\hat{s}$}. 
Then, for each class $s$, we extract the features of all its unlabeled samples by $f$ and initialize our prototypes $p_j^s$ (Fig.~\ref{clustering_figure}, black marks) as the $k$-means cluster centers of the extracted features.
In this step, both $K$ and $M$ are hyperparameters.

\vspace*{2mm}
\noindent {\bf Clustering Loss.} 
For each unlabeled sample $\hat{x}_i$ in a mini-batch during training, given its pseudo label \textcolor{myblue}{$\hat{s}$} (generated by SSL method) and the associated $K$ prototypes $p^s_j, j \in \{1,2,...,K\}$ of class $s$, we define our prototype-based clustering loss as:
\begin{equation}
\begin{aligned}
   L^s_{c}(\hat{x}_i) &= \\ - &\log {\frac{
      \mathrm{exp}(f(\hat{x}_i) \cdot p_*^s/\tau)
      }
      {
         \sum^{K}_{j=1}{
            \mathrm{exp}(f(\hat{x}_i) \cdot p_j^s/\tau)
      }
   }
   \mathbbm{1}(\max(F(\hat{x}_i)) > t_c)},
   \label{equation_L_cluster}
\end{aligned}
\end{equation}
where $\mathbbm{1}$ is an indicator function, $t_c$ is a threshold parameter, 
$p_*^s$ is the prototype that is closest to $f(\hat{x}_i)$ in Euclidean space,
$\tau$ is a temperature parameter used in self-supervised learning~\cite{chen2020simple}.
Intuitively, minimizing $ L^s_{c}(\hat{x}_i)$ guides classifier $F$ to generate features $f(\hat{x}_i)$ closer to $p^s_*$ but further to other prototypes.
Thus, the overall loss across a mini-batch of unlabeled samples is:
\begin{equation}
   L_{c} = \sum_{j=1}^{BS}  L^s_{c}(\hat{x}_j)
   \label{eq:L_cluster_bath}
\end{equation}
where $BS$ is the batch size.
$L_{c}$ is used as an additional loss term in the SSL loss function.

\textcolor{myblue}{In addition to the clustering loss for unlabeled data, we further apply a similar loss to the labeled data, clustering them to the same centers.
% which forces the labeled samples to be clustered to the same center. 
% Specifically, given a labeled sample $x_i$ and its ground truth label $y$, we define the loss as:
Specifically, given a labeled sample $x_i$ ($1 \leq i \leq N_y$) and its ground truth label $y$, where $N_y$ is the total number of labeled samples with label $y$, we define the loss as:
\begin{equation}
    L_{labeled}^{y}(x_i) = - f(x_i) \cdot q_y 
 + L_c^s(x_i),
\label{eq:clustering_labeled}
\end{equation}
where $q_y$ is the normalized feature center of class $y$:
\begin{equation}
    q_y = \mathrm{normalize}(\sum_{i=1}^{N_y}f(x_i)).
\end{equation}
% where $N_y$ is the total number of labeled samples with label $y$.
% \begin{equation}
%     q_y = \mathrm{normalize}(\sum_{j=1}^{N_y}f(x_j)).
% \end{equation}
% $N_y$ is the number of labeled samples of class $y$.
In Eq.~\ref{eq:clustering_labeled}, the first term aims to cluster all labeled samples of the same class towards their normalized feature center $q_y$, preventing them from being misaligned to different ID/OOD centers; the second term applies the same clustering loss for unlabeled data (Eq.~\ref{equation_L_cluster}) to labeled data, indicating that both the labeled and unlabeled ID samples should be clustered to the same centers. 
% The first term of $L_{labeled}^{y}(x_i)$ aims to cluster the labeled samples of the same class. 
% Note that the second term is the same loss for unlabeled data, which indicates that the labeled samples should be aligned to the same cluster as unlabeled data. 
% Simultaneously minimizing both terms will clustered the labeled samples to the same center and avoid the separation of labeled samples to different ID/OOD centers.
The clustering loss clusters the features of both labeled and unlabeled data in a semi-supervised manner, thus separates the ID and OOD data into different cluster centers.}

\vspace*{2mm}
\noindent {\bf Prototype Update.}
At the same time, for each sample $\hat{x}_i$ in a mini-batch during training, we dynamically update its nearest prototype $p^s_*$ as a moving average:
\begin{equation}
  p_{*,(t+1)}^{s} = \alpha p_{*,(t)}^{s} + \beta f(\hat{x}_i),
  \label{equation_prototype_update}
\end{equation}
where $\alpha=0.99$ and $\beta=0.01$ are weighting parameters 
\textcolor{myblue}{following the common practice of momentum update. }

\vspace*{2mm}
With our prototype-based clustering algorithm, pseudo-labeled samples in each class $s$ are clustered according to the similarity of their features.
As a result, heterogeneous samples are pushed away from each other.
This helps SSL as the difference between OOD and ID samples are also clarified, thereby helping the feature extractor to learn better representations.
Thus, OOD samples are less likely to be misclassified as ID samples and damage the self-training.
Based on the clustering results, we identify ID/OOD samples as follows.
%%%%%%%%%%%%%%%%
%下面加入一些针对exploiting OOD data为什么work的解释
%%%%%%%%%%%%%%%%
% Besides, prototype-based clustering algorithm helps the feature extractor organize both ID and OOD features and thus improve the representation.

\vspace*{2mm}
\noindent {\bf Sample Identification.}
First, we identify the {\it unlabeled} ID samples to be included in the pools according to their distances to the {\it labelled} ID samples in the labeled dataset $\mathcal{X}=\{(x_i,s_i)\}_{i=1}^{N_l}$, $s_i \in \{1,2,...,S\}$ where $S$ is the number of classes.
Let $f(x_i)$ be the \textcolor{myblue}{normalized} feature of $x_i$ that is extracted by $f$, we can calculate the per-class feature centers of the labeled data as:
\begin{equation}
   O_s = \frac{\sum_{i=1}^{N_l} \mathbbm{1}(s_i = s) f(x_i)}{\sum_{i=1}^{N_l} \mathbbm{1}(s_i = s)}, s \in \{1,2,...,S\}
\end{equation} 
where $\mathbbm{1}$ is an indicator function. Since all labeled data are ID, an unlabeled sample with pseudo label $s$ tend to be ID if its corresponding prototype $p_*^s$ is close to $O_s$. Therefore, for each class $s$, we compute the Euclidean distances from $O_s$ to each of its prototypes $p^s_j$. 
According to these distances, we sort all prototypes $p^s_j$ in increasing order and pick the first $N_{id}$ of them as ID prototypes. 
For each unlabeled sample $\hat{x_i}$ in a mini-batch, we identify it as ID if its closest prototype $p_*^s$ is an ID prototype.
Otherwise, $\hat{x_i}$ is identified as OOD.
In this step, $N_{id}$ is a hyperparameter.

\subsection{Importance Sampling for Open-Set SSL}
\label{sec:sample_refinement}

% Recalling the definition of Open-set SSL where the dataset contains two types of samples, \ie~ID and OOD samples, it is straightforward to assume that they are of different importance to SSL: ID samples are useful while OOD ones are irrelevant. 
\textcolor{myblue}{Recalling the definition of Open-set SSL where the dataset contains two types of samples, \ie~ID and OOD samples, it is straightforward to assume that they are of different importance to SSL: ID samples are useful while OOD ones are less relevant for the target task. }
Such an assumption motivates the selection of ID samples in SSL that is widely employed in previous methods.
\textcolor{myblue}{Despite that OOD samples can help the training of semi-supervised methods, overwhelming OOD samples in mini-batches will unstabilize the training and lower the performance when the images are large and the batch size is small.}
% We follow this paradigm and also use ID samples to train our classifier.
\textcolor{myblue}{Therefore, We follow the ID sample selection paradigm and try to improve the performance by making the classifier concentrate on ID samples while utilizing the information of OOD samples.}
% However, unlike previous methods, we noticed that ID samples are dynamically identified during training. 
Unlike previous methods, we noticed that ID samples are dynamically identified during training.
Thus, the ID samples identified earlier occurs more often in random sampling and are thus biased. 
This is undesirable as they can soon be well-learnt and contribute less to the training than newly identified ones.
To this end, we propose a novel importance sampling method \textcolor{myblue}{for mini-batch sampling during training} that assigns importance scores to unlabeled samples and only maintains the important ones in the sample pools as follows.

\vspace*{2mm}
\noindent {\bf Importance-based Sample Pools.}
After identification, the identified OOD samples are assigned importance scores of 0 and excluded from SSL; the identified ID samples are assigned importance scores of $1/I(\hat{x}_i)$ where $I(\hat{x}_i)$ is the number of times $\hat{x}_i$ is identified as an ID sample throughout the training. We maintain the identified ID samples in our per-class importance-based sample pools $P_s$, where $s\in \{1,2,...,S\}$ is the class label.
We restrict $P_s$ to contain at most $N_p$ samples, $N_p \gg BS$ where $BS$ is the batch size. 
% During mini-batch training, assume that $M$ ID samples are identified for class $s$ in one iteration, we update $P_s$ by:
During mini-batch training, assume that $M$ ID samples are identified for class $s$ in one iteration, we update $P_s$ by:
% \begin{itemize}
    
    % \item {\bf Case 1.} 
    \noindent \emph{- Case 1.}
    If $P_s$ has enough space, we simply add the $M$ ID samples to $P_s$.
    
    % \item {\bf Case 2.} 
    \noindent \emph{- Case 2.}
    % Otherwise, we assign a probability (importance score) $\mathcal{P}_{\hat{x}_i}$ for each sample $\hat{x}_i$ in $P_s$, $i \in \{1,2,...,N_{P}\}$:
    Otherwise, we compute probability $\mathcal{P}_{\hat{x}_i}$ for each sample $\hat{x}_i$ in $P_s$, $i \in \{1,2,...,N_{P}\}$ using their importance scores as:
    % Otherwise, we devise a hard-sample-mining mechanism for $P_s$ as follows. First, we assign a probability $\mathcal{P}_{\hat{x}_i}$ for each sample $\hat{x}_i$ in $P_s$, $i \in \{1,2,...,N_{P}\}$:
    %\item {\bf Case 2.} Otherwise, we devise a self-correction mechanism for $P_s$ as follows. First, we assign a probability f$\mathcal{P}_{\hat{x}_i}$ for each sample $\hat{x}_i$ in $P_s$, $i \in \{1,2,...,N_{P}\}$:
    \iffalse
    \begin{equation}
        \mathcal{P}_{\hat{x}_i} = \min\{(N_p - M) \frac{I(\hat{x}_i)} {\sum_{j=1}^{N_p}I(\hat{x}_j)}, 1\},
   \label{p_i_replacement}
   \end{equation}
   \fi
   \begin{equation}
        \mathcal{P}_{\hat{x}_i} = \min\{ M \frac{I(\hat{x}_i)} {\sum_{j=1}^{N_p}I(\hat{x}_j)}, 1\},
   \label{p_i_replacement}
    \end{equation}
    % where $I(\hat{x}_i)$ is the number of times $\hat{x}_i$ is identified as an ID sample.
    %Note that $\mathcal{P}_{\hat{x}_i} \ge 0$ because $M \leq BS \ll N_p$.
    Then, we select each sample by probability $\mathcal{P}_{\hat{x}_i}$ and obtain $N_{r}$ samples.
    We replace the first $\min (N_{r}, M)$ of them with the newly identified ID samples. 
    % Intuitively, an unlabeled sample is more likely to be removed from $P_s$ if it is always identified as ID, i.e., it is well-learnt.
    Intuitively, an ID sample is more likely to be removed from $P_s$ if it is sampled more often, i.e., it is well-learnt.
    % In this way, we keep $P_s$ healthy by storing the relatively difficult ID samples.
% \end{itemize}

% However, we empirically observed that the accuracy of ID sample identification is not high enough. 
 % result, the potential oour ID sample pools is not fully utilized as they still contain many misclassified ID samples.
% To this end, we devise a cascading pooling strategy to further improve the density of ID samples as follows.
\textcolor{myblue}{However, it is difficult to identify ID samples accurately by performing the identification once when the unlabeled set is complicated and OOD data can benefit the semi-supervised training. Besides, the density of ID samples in a mini-batch is important for the performance as we show in Sec~\ref{mask_ood_section}. To this end, we devise a cascading pooling strategy to further improve the density of ID samples as follows, and it can help and stabilize the SSL training by providing high-density ID samples within a mini-batch.}

\vspace*{2mm}
\noindent {\bf Cascading Sample Pools.}
% \noindent {\bf Level-wise Sample Pools.}
Let $S$ be the number of classes, we cascade different sets of sample pools as a pyramid:
\begin{itemize}
    \item Level 0 of the pyramid is the raw dataset.
    \item Level 1 is a set of $S$ ID sample pools. The capacity of each sample pool is $N_P$.
    \item Level 2 is a set of $S$ ID sample pools. The capacity of each sample pool is $N_P / 2$.
    \item ......
    \item Level N is a set of $S$ ID sample pools. The capacity of each sample pool is $N_P / 2^{N-1}$.
\end{itemize}
During training, we circularly draw mini-batches of samples in a level-wise manner from Level 0 to Level N.
In each training iteration, we draw samples evenly from the $S$ sample pools in the same level and apply ID sample identification to it. The newly identified ID samples are used to update the sample pools at the next level.
% Empirically, we verified the effectiveness of this strategy in Sec.~\ref{id_density_section}.

\section{Experiment}

\subsection{Experimental Setup}
\label{sec:experimental_setup}

\vspace*{2mm}
\noindent {\bf Datasets.}
Following the common practice in SSL evaluation~\cite{sohn2020fixmatch}, we test our method on \textcolor{myblue}{four} benchmark datasets:

% \begin{itemize}
    % \item {\bf CIFAR-100~\cite{krizhevsky2009learning}}: 
    \noindent \emph{- CIFAR-100~\cite{krizhevsky2009learning}}: 
    a dataset consisting of 100 classes of natural images. Each class contains 500 training images and 100 testing images.
    
    \noindent \emph{- SVHN~\cite{netzer2011reading}}: 
    a dataset consisting of 10 classes of digits images. It contains 73,257 and 26,032 digits images for training and testing respectively.
    
    % \item {\bf TinyImageNet}:
    \noindent \emph{- TinyImageNet}:
    a subset of the ImageNet dataset~\cite{deng2009imagenet} consisting of 200 classes of natural images. Each class contains 500 training images and 50 test images.
% \end{itemize}

And a more challenging and realistic dataset:
% \begin{itemize}
    
    % \item {\bf DomainNet-Real~\cite{peng2019moment}}:
    \noindent \emph{- DomainNet-Real~\cite{peng2019moment}}:
    DomainNet is a dataset consisting of 345 classes of images in 6 domains ({\it e.g.} real, painting, sketch). In our experiments, we only use the 172,947 images in its Real domain as we observed that FixMatch~\cite{sohn2020fixmatch} performs poorly in some domains.
% \end{itemize}

\begin{table*}

    \caption{Experimental results on Open-Set Semi-Supervised Learning. Our method outperforms MTCF~\cite{yu2020multi} and improves FixMatch~\cite{sohn2020fixmatch} by a significant margin. ID / OOD: the number of classes whose images are defined as ID and OOD samples respectively. $(\cdot)$ / 50k$^*$: to balance the numbers of ID and OOD samples, we sample 50k images from the classes other than $(\cdot)$ in DomainNet-Real as OOD samples.
    $^\dagger$: DS$^3$L consumes too much memory and time on DomainNet-Real and cannot run on commodity workstations.}
    \label{tab:main}
   \begin{center}
   {\color{myblue}\begin{tabular}{l c c c c c c}
   \toprule
   Datasets & \multicolumn{2}{c}{DomainNet-Real} & \multicolumn{2}{c}{CIFAR-100} & \multicolumn{2}{c}{TinyImageNet} \\
   \midrule
   ID / OOD & 10 / 50k$^*$ & 20 / 50k$^*$ & 10 / 90 & 20 / 80 & 10 / 190 & 20 / 180 \\
   \midrule
   Labeled Only & 48.5$\pm$1.0 & 41.6$\pm$0.7 & 47.3$\pm$1.8 & 40.0$\pm$0.4 & 36.9$\pm$2.3& 32.2$\pm$0.9 \\
   
   FixMatch~\cite{sohn2020fixmatch} & 52.8$\pm$2.9 & 49.7$\pm$2.5 & 80.8$\pm$0.9 & 72.2$\pm$0.2 & 68.9$\pm$0.7 & 53.6$\pm$1.0 \\
   MTCF~\cite{yu2020multi} & 54.2$\pm$1.8 & 46.3$\pm$0.4 & 59.8$\pm$0.6 & 46.2$\pm$1.0 & 52.4$\pm$1.2 & 46.5$\pm$0.6 \\
   DS$^3$L\cite{guo2020safe}$^\dagger$ & -- & -- & 57.0$\pm$0.7 & 40.2$\pm$1.0 & 52.2$\pm$2.7 & 40.0$\pm$1.6    \\
   Energy\cite{liu2020energy} & 50.1$\pm$1.8 & 45.9$\pm$1.0 & 82.5$\pm$0.7 & 72.9$\pm$1.6 & 67.3$\pm$2.0 & 56.5$\pm$1.5 \\
   ReAct\cite{sun2021react} & 50.1$\pm$1.1 & 46.6$\pm$0.7 & 82.9$\pm$0.7 & 73.3$\pm$2.0 & 69.5$\pm$1.7 &  57.7$\pm$2.0 \\
   OpenMatch\cite{saito2021openmatch} & 54.8$\pm$2.6 & 50.4$\pm$1.2 & 83.0$\pm$1.0 & 73.3$\pm$2.5 & 68.7$\pm$2.8 & 54.8$\pm$1.0 \\
   Ours & \textbf{59.4$\pm$0.3} & \textbf{54.3$\pm$1.2} & \textbf{85.5$\pm$0.8} &  \textbf{76.0$\pm$1.1} & \textbf{71.4$\pm$0.7} & \textbf{58.5$\pm$1.1} \\
   \midrule
   Clean & 63.5$\pm$0.7 & 60.7$\pm$0.8 & 84.8$\pm$0.7 & 72.3$\pm$0.4 & 79.5$\pm$0.8 & 60.3$\pm$0.3 \\
   \bottomrule
   \end{tabular}}
   \end{center}
\end{table*}

\vspace*{2mm}
\noindent {\bf Implementation Details.}
We implement our method on top of FixMatch~\cite{sohn2020fixmatch}, a state-of-the-art SSL algorithm.
In addition to the relatively standard pseudo labeling, FixMatch used another common SSL technique: consistency regularization. In a nutshell, it encourages the SSL classifier to output the same value for two variants of an unlabeled sample $\hat{x}_i$: a weak-augmented variant $\hat{x}^a_i$ and a strong-augmented variant $\hat{x}^b_i$.
Accordingly, for class $s$ and sample $\hat{x}_i$, we extend $L^s_{c}$ (Eq.~\ref{equation_L_cluster}) to $L_{c}^{s,\mathrm{CR}}$ as follows:
\begin{equation}
  L_{c}^{s,\mathrm{CR}}(\hat{x}_i) = L^s_{c}(\hat{x}^a_i) + L^s_{c}(\hat{x}^b_i)
\end{equation}
Note that we use the same target prototype $p_*^s$ that is closest to the weak-augmented variant $\hat{x}^a_i$ for both $L^s_{c}(\hat{x}^a_i)$ and $L^s_{c}(\hat{x}^b_i)$ because i) heuristically, $\hat{x}^a_i$ is weak-augmented and thus closer to $\hat{x}_i$ in the feature space; ii) in line with consistency regularization, $\hat{x}^a_i$ and $\hat{x}^b_i$ share the same semantic meanings and should be in the same cluster, {\it i.e.} with the same prototype. Similarly, we only use the weak-augmented variants of unlabeled samples in prototype update and sample identification.
Following FixMatch, we employ different network architectures for different datasets. \textcolor{myblue}{We tune the hyper-parameters using a small validation set.}
\begin{itemize}
    \item For CIFAR-100, SVHN and TinyImageNet, we follow FixMatch~\cite{sohn2020fixmatch} and use the same architecture based on Wide ResNet (WRN 28$\times$8)~\cite{zagoruyko2016wide}. All images are resized to 32 $\times$ 32. 
    % We set the number of prototypes $K=10$ and the weight of $L_c^s({\hat{x_i}})$ is 0.01.
    We set the number of prototypes $K=10$ and the weight of $L_c$ as 0.01 when added to the FixMatch loss function.
    % Following~\cite{chen2020simple}, we set $\tau=0.07$ and $t_c=0.98$, which is slightly higher than the FixMatch threshold 0.95 (to select sample for FixMatch training).
    Following~\cite{chen2020simple}, we set $\tau=0.07$ and $t_c=0.98$, which is slightly higher than FixMatch's pseudo labeling threshold of 0.95. We use the same hyperparameters of FixMatch~\cite{sohn2020fixmatch} in the semi-supervised learning (SSL) part of our method. 
    We run our method on 1 Nvidia Tesla V100 GPU with 16GB memory and set the batch size as 64 for labeled data and 448 (64$\times$7) for unlabeled data. 
    We report the experimental results after 100 epochs of training. 
    \item For DomainNet-Real, we use the ResNet-50~\cite{he2016deep} architecture. All images are resized to 224$\times$224. 
    % Following the ImageNet~\cite{deng2009imagenet} training scheme in FixMatch~\cite{sohn2020fixmatch}, we set $t_c=0.7$ which equals the FixMatch threshold. $K$ is set as 30 and clustering loss weight is 0.1. Additional details are provide in the supplementary material.
    We set the number of prototypes $K=30$ and the weight of $L_c$ as 0.1 when added to the FixMatch loss function.
    Following the ImageNet~\cite{deng2009imagenet} training scheme in FixMatch~\cite{sohn2020fixmatch}, we set $t_c=0.7$ which equals to FixMatch's pseudo labeling threshold. In our experiment, $N_{id} = K/5$. 
    We use the same hyperparameters of FixMatch~\cite{sohn2020fixmatch}.
    We run our method on 6 Nvidia Tesla V100 GPUs and report the experimental results after 100 epochs of training. 
    For each GPU, we set the batch size as 8 for labeled data and 56 for unlabeled data. 
    Following FixMatch~\cite{sohn2020fixmatch}, we apply linear warmup to the learning rate for the first 5 epochs of training until it reaches an initial value of 0.4. 
    % Then, we decay the learning rate at epoch 60 by multiplying it by 0.1. 
    At epoch 60, we decay the learning rate by multiplying it by 0.1.
\end{itemize}

\vspace*{2mm}
\noindent {\bf Experimental Settings.}
(1) ID {\it vs.} OOD. For CIFAR-100, TinyImageNet and SVHN, {\it ID samples} are defined as the images in the first $N$ classes; {\it OOD samples} are defined as those in the rest classes.
For DomainNet-Real, {\it ID samples} are defined as the images in the $N$ classes with the most images; {\it OOD samples} are defined as the 50k images sampled from the rest classes, which aims to balance the numbers of ID and OOD samples.
(2) Labelled {\it vs.} unlabelled. For all datasets, {\it labelled data} are defined as the first 25 images and their associated labels in each of the $N$ classes; {\it unlabelled data} are defined as the rest images in each of the $N$ classes together with the OOD samples.
(3) Training {\it vs.} Testing. For DomainNet, for each of the $N$ classes, the {\it testing set} is defined as the $100$ images sampled from the unlabelled data in the class; For other three datasets we directly use the pre-defined testing set; the {\it training set} is defined as other images (including both labelled and unlabelled data) in the class.
Furthermore, we report our method's average performance of the last 10 epochs over 3 runs using the same random seed set. 

\begin{table*}
\caption{Justification of OOD samples' benefits in SSL. ID / OOD: the number of classes whose images are defined as ID and OOD samples. S10 / C100$^*$: 10 ID classes are selected from SVHN and 100 OOD classes are selected from CIFAR-100.}
\begin{subtable}[h]{0.4\textwidth}
   \begin{center}
   \begin{tabular}{l c}
   \toprule
   Method & DomainNet-Real \\
   \midrule
   FixMatch~\cite{sohn2020fixmatch} & 49.7$\pm$2.5 \\
   Mask-OOD & 54.3$\pm$0.7 \\
   SimCLR-OOD & 56.7$\pm$0.4 \\
%   Ours & 53.4$\pm$1.5\\
%   \midrule
   Clean & 60.7$\pm$0.8\\
   \bottomrule
   \end{tabular}
   \end{center}
    \subcaption{Normal case}
   \label{ood_ablation}
\end{subtable}
\hfill
\begin{subtable}[h]{0.6\textwidth}
   \begin{center}
   \begin{tabular}{l c c c}
   \toprule
   Datasets & CIFAR-100 & SVHN & TinyImageNet \\
   \midrule
   ID / OOD   & 10 / 90  & S10 / C100$^*$ &  20 / 180 \\
   \midrule
   Labeled Only      & 47.3$\pm$1.8    & 24.6$\pm$2.4   & 32.2$\pm$0.9         \\
   FixMatch          & 68.7$\pm$1.5   & 43.4$\pm$2.7   & 46.9$\pm$0.4         \\
   Ours (Clustering) & \textbf{73.5$\pm$1.3}     & \textbf{50.2$\pm$2.9}   & \textbf{52.0$\pm$0.9}         \\
%   \midrule
%   Clean            & 84.8$\pm$0.7       & 90.7$\pm$0.1       &     60.3$\pm$0.3    \\    
   \hline
   \end{tabular}
   \end{center}
   \subcaption{Extreme case}
   \label{OOD_only_table}
\end{subtable}

\end{table*}

\subsection{Experimental Result}
\label{sec:experimental_result}

% As Table~\ref{tab:main} shows, our method significantly outperforms both MTCF~\cite{yu2020multi} and FixMatch~\cite{sohn2020fixmatch}.
\textcolor{myblue}{As Table~\ref{tab:main} shows, our method significantly outperforms previous open-set semi-supervised learning and OOD detection methods including MTCF~\cite{yu2020multi}, DS$^3$L\cite{guo2020safe}, Energy\cite{liu2020energy}, ReAct\cite{sun2021react} and OpenMatch\cite{saito2021openmatch}.}
To give a better idea on how good our method performs, we provide two additional baselines using FixMatch~\cite{sohn2020fixmatch}:
\begin{itemize}
\item ``Labeled Only'': a FixMatch model trained with labeled data only, which can be viewed as a lower bound.
\item ``Clean'': a FixMatch model trained with ID samples only, which can be viewed as an improved baseline.
\end{itemize}
We test all methods on three datasets: CIFAR-100, TinyImageNet and DomainNet-Real\footnote{We did not use SVHN because it has only 10 classes and thus cannot fit into this experiment.}.
% As discussed in Sec.~\ref{sec:experimental_setup}, for each dataset, we define the images in its first $N$  ($N=10,20$) classes as the ID samples; the OOD samples are defined accordingly.
As discussed in ``Experimental Setup'' section, for each dataset, we define the images in its first $N$  ($N=10,20$) classes as the ID samples; the OOD samples are defined accordingly.
\begin{itemize}
    \item For CIFAR-100 and TinyImageNet, we observed small gaps between FixMatch and Clean, which leaves small room for improvement.
    Similar to~\cite{yu2020multi}, we conjecture that the reason is the relatively simple datasets being used.
    However, it is interesting to see that our method outperforms ``Clean'' on CIFAR-100, 10/90 and 20/80 (10/20 ID classes and 90/80 OOD classes from CIFAR-100). This implies that OOD samples are also useful in SSL, which contradicts the common belief that OOD samples are harmful. 
    Nevertheless, we propose to test on a more realistic and challenging dataset: DomainNet-Real.
    \item For DomainNet-Real, we observed approximately $10\%$ gaps between FixMatch and Clean. In such challenging scenarios, our method also significantly outperforms MTCF~\cite{yu2020multi} and FixMatch~\cite{sohn2020fixmatch}. However, there is a considerable gap between our method and ``Clean'', which suggests that there is still room for improvement.
\end{itemize}

In summary, experimental results show that our method performs the best against competing methods in all six settings (two for each dataset), which indicates that the improvement brought by our method can be generalized to a variety of datasets and ID/OOD ratios.

\subsection{Do OOD Samples Really Benefit SSL?}
\label{mask_ood_section}

This section justifies the motivation of our method: if being ``properly'' used, OOD samples can benefit SSL.
To verify this claim, we assume that {\it all unlabelled samples are perfectly identified as ID and OOD samples before training}.
Base on this assumption, we propose two strategies to handle the OOD samples when training a FixMatch~\cite{sohn2020fixmatch} SSL model:
\begin{itemize}
    \item {\bf Mask-OOD} masks all OOD samples by setting their weights to 0 in the FixMatch loss function.
    \item {\bf SimCLR-OOD} adds a SimCLR loss term~\cite{chen2020simple} for OOD samples in the FixMatch loss function. 
\end{itemize}
Table~\ref{ood_ablation} shows the results of Mask-OOD and SimCLR-OOD against the original FixMatch and ``Clean'' on the DomainNet-Real dataset. 
We use the same hyperparameters for all methods.
Note that Mask-OOD is different from ``Clean'' as it does not remove the OOD samples and thus keeps the density of ID samples in mini-batches.
It can be observed that: 
i) Mask-OOD works better than the original FixMatch, which is consistent with the common belief that OOD is harmful to SSL.
ii) Mask-OOD works worse than SimCLR-OOD, which justifies our claim that {\it compared to filtering out OOD samples, exploiting them properly benefits SSL.}
iii) Mask-OOD works worse than ``Clean'', which indicates that the performance of SSL depends on the density of ID samples in a mini-batch. 
This motivates the use of our cascading pooling strategy.

\vspace*{2mm}
\noindent {\bf Extreme Case Study.}
To further justify the motivation of our method, we test the performance of our method in an extreme case of open-set SSL where {\it all unlabeled samples are OOD}.
To implement it, we remove all unlabeled ID samples from the training dataset.
% Note that we also remove the sample refinement strategy from our method as it is useless in this scenario.
Note that we also remove the importance-based sampling method as it is useless in this scenario. 
% In this section, we ablate our sample refinement and justify the effectiveness of our prototype-based clustering algorithm.
% To eliminate the usefulness of sample refinement, we assume all unlabeled samples are OOD, which is an extreme case of open-set SSL.
Specifically, we compare our method (with clustering only) with FixMatch~\cite{sohn2020fixmatch} and its variant ``Labeled Only''\footnote{In this case, ``Clean'' degenerates to ``Labeled Only''.} on three datasets: CIFAR-100, SVHN and TinyImageNet.
% Specifically, we compare our method (with clustering only) with FixMatch~\cite{sohn2020fixmatch} and its variant ``Labeled Only''\footnote{In this case, ``Clean'' degenerates to ``Labeled Only''.} (Sec.~\ref{sec:experimental_result}) on three datasets: CIFAR-100, SVHN and TinyImageNet.
For CIFAR-100 and TinyImageNet, we set up the labeled ID samples and the unlabelled OOD samples within the same datasets.
For SVHN, we set up the labeled ID samples from all its 10 classes and borrow the images from CIFAR-100 as the unlabelled OOD samples.
As Table~\ref{OOD_only_table} shows, it can be concluded that
% i) Our prototype-based clustering algorithm is effective because Ours (Clustering) outperforms the original FixMatch.
Unlabeled OOD samples can still benefit SSL without unlabelled ID samples, which is justified by the observation that both Ours (Clustering) and FixMatch outperform ``Labeled Only''. 
% This further justifies the motivation of our method that OOD samples {\it DO} benefit SSL.
This further justifies our motivation that OOD samples {\it DO} benefit SSL.

\begin{table}
\caption{Ablation study. $(\cdot)$ / 50k$^*$: to balance the numbers of ID and OOD samples, we sample 50k images from the classes other than $(\cdot)$ in DomainNet-Real as OOD samples.}
   \begin{center}
   {\color{myblue}\begin{tabular}{l c c}
   \toprule
   Datasets & \multicolumn{2}{c}{DomainNet-Real} \\
   \midrule
   Method & 10/50k$^*$ & 20/50k$^*$ \\
   \midrule
   FixMatch~\cite{sohn2020fixmatch} & 52.8$\pm$2.9 & 49.7$\pm$2.5 \\
   + clustering (Weak-Aug-Only) & 54.2$\pm$1.4 & 51.6$\pm$0.5 \\
   + clustering & 55.0$\pm$1.1 & 52.5$\pm$0.7 \\
   + refinement (Random) & 57.2$\pm$1.2 & 53.1$\pm$1.3\\
   + refinement (Importance) & 58.1$\pm$0.4 & 53.7$\pm$0.8\\
   + refinement (Ours) & 59.4$\pm$0.3 & 54.3$\pm$1.2\\
   \midrule
   Clean & 63.5$\pm$0.7 & 60.7$\pm$0.8\\
   \bottomrule
   \end{tabular}}
   \end{center}
%   \caption{Ablation study on clustering and refinement in DomainNet-Real with two different ID/OOD settings.}

   \label{overall_ablation}
\end{table}

\subsection{Ablation Study}
\label{id_density_section}

This section studies the extent to which our proposed prototype-based clustering and identification algorithm and our importance sampling method contribute to the performance gains respectively.
% This section studies how much our proposed prototype-based clustering algorithm and sample refinement strategy contribute to the performance gain respectively.
Specifically, we start from the original FixMatch~\cite{sohn2020fixmatch} and add our prototype-based clustering and identification algorithm and our importance-based sampling method in turn.
% Besides, we apply different variants of our two components to further demonstrate the effectiveness of our method. 
\textcolor{myblue}{To further demonstrate the effectiveness of our method, we also tested several variants of our two components, including:
{\it Clustering (Weak-Aug-Only)}, which ignores the strongly-augmented samples and only clusters the weakly-augmented samples during training; 
{\it Refinement (Random)}, which randomly selects the ID samples identified in the clustering procedure for training;
{\it Refinement (Importance)}, which removes the cascading sample pools and only uses importance-based sampling for training. }
% Clustering (Weak-Aug-Only) only clusters the weakly-augmented samples and ignores the strongly-augmented samples during training. 
% Refinement (Random) randomly selects the samples that are classified as in-distribution in the clustering procedure for training.
% Refinement (Importance) uses importance-based sampling for training without cascading pooling. 
% Specifically, we start from the original FixMatch~\cite{sohn2020fixmatch} and add our prototype-based clustering and identification algorithm and sample refinement strategy in turn.
All these methods are tested on two settings of the DomainNet-Real dataset with ``Clean'' as a reference.
The experimental results are shown in Table~\ref{overall_ablation}.
It can be observed that:
i) Our clustering and identification algorithm improves the performance over FixMatch by 2.2\% and 2.8\% respectively. 
% i) Our clustering algorithm improves the performance over FixMatch by 1.5\% and 2.3\% respectively.
ii) Adding our importance-based sampling method can further improve the performance by 4.4\% and 1.8\% respectively (\ie 6.6\% and 4.6\% higher than FixMatch).
% ii) Adding our sample refinement strategy can further improve the performance by 3.8\% and 1.4\% respectively (\ie 5.7\% and 3.7\% higher than FixMatch).
Note that ``importance sampling only'' is not a valid variant because our importance-based sampling method relies on the identification results and cannot be used independently.
% Note that ``refinement only'' is not a valid variant because our sample refinement strategy relies on the result of our clustering algorithm and cannot be used independently.

\iffalse
To further justify the effectiveness of our cascading pooling strategy, we plot the density of ID samples in our two cascaded ID sample pools (with per-class capacity 300 and 150 respectively) against training epochs when training our model with the DomainNet-Real 20/50k setting (Figure~\ref{id_density_figure}).
% To further justify the effectiveness of our sample refinement strategy, we plot the the density of ID samples in our two cascaded ID sample pools (with per-class capacity 300 and 150 respectively) against training epochs when training our model with the DomainNet-Real 20/50k setting (Figure~\ref{id_density_figure}).
We also marked the percentage of ID sample in the raw unlabeled dataset as ``Random Selection''. 
It can be observed that:
i) Our two pools have much higher ID sample densities than Random Selection (approximately 10\% and 20\% respectively), which justifies the usefulness of our approach.
ii) Pool 2 has a much higher ID sample density than Pool 1 (approximately 10\%), which indicates the effectiveness of our cascading pooling strategy.
\fi

To demonstrate that our method generalizes to other SSL methods, we integrate our method to UDA~\cite{xie2019unsupervised} and FlexMatch\cite{zhang2021flexmatch} and test their performance on CIFAR-100 dataset (Table \ref{tab:UDA}).
% To investigate the performance of our method on other SSL model, here we adopt UDA\cite{xie2019unsupervised} as SSL baseline\footnote{https://github.com/ildoonet/unsupervised-data-augmentation} and use CIFAR100 in experiments. 
% We report the experimental results after 1600 epochs of training.
It can be observed that our method improves the performance of UDA and FlexMatch under open-set settings by a significant margin.

\begin{table}
    \caption{\textcolor{myblue}{Integrating our method in UDA~\cite{xie2019unsupervised} improves its performance on open-set SSL tasks (CIFAR-100).}}
       \begin{center}
       {\color{myblue}\begin{tabular}{l c c}
       \toprule
       Datasets & \multicolumn{2}{c}{CIFAR-100} \\
       \midrule
       ID/OOD & 10/90 & 20/80 \\
       \midrule
       UDA~\cite{xie2019unsupervised} & 38.9$\pm$1.5 &  39.9$\pm$2.1 \\
       + Our Method & 48.4$\pm$1.1 & 43.1$\pm$1.7 \\
       \midrule
       Clean(UDA) & 67.9$\pm$0.5 & 63.4$\pm$0.9 \\
       \midrule
       FlexMatch~\cite{zhang2021flexmatch} & 86.6$\pm$0.3  & 80.9$\pm$0.8  \\
       + Our Method & 88.0$\pm$0.3 & 84.8$\pm$0.4 \\
       \midrule
       Clean(FlexMatch) & 88.1$\pm$0.1 & 83.1$\pm$0.1 \\
       \bottomrule
       \end{tabular}}
       \end{center}
       \label{tab:UDA}
\end{table}

\begin{table}
    % \caption{The comparison of ID/OOD classification AUROC results.}
    \caption{\textcolor{myblue}{Comparison of AUROC values for ID/OOD classification.}}
       \begin{center}
       {\color{myblue}\begin{tabular}{l c c}
       \toprule
       Datasets & \multicolumn{2}{c}{CIFAR-100} \\
       \midrule
       ID/OOD & 10/90 & 20/80 \\
       \midrule
       FixMatch & 60.2$\pm$0.2 &  57.0$\pm$0.3 \\
       MTCF & 70.6$\pm$1.1 &  68.9$\pm$1.4 \\
       OpenMatch & 72.3$\pm$0.8 &  71.5$\pm$0.2 \\
       \midrule
       Our Method & 79.6$\pm$0.7 & 73.5$\pm$0.5 \\
       \bottomrule
       \end{tabular}}
       \end{center}
       \label{tab:auroc}
\end{table}

\begin{figure}[!htb]
    \begin{center}
        \includegraphics[height=0.5\linewidth,width=0.8\linewidth]{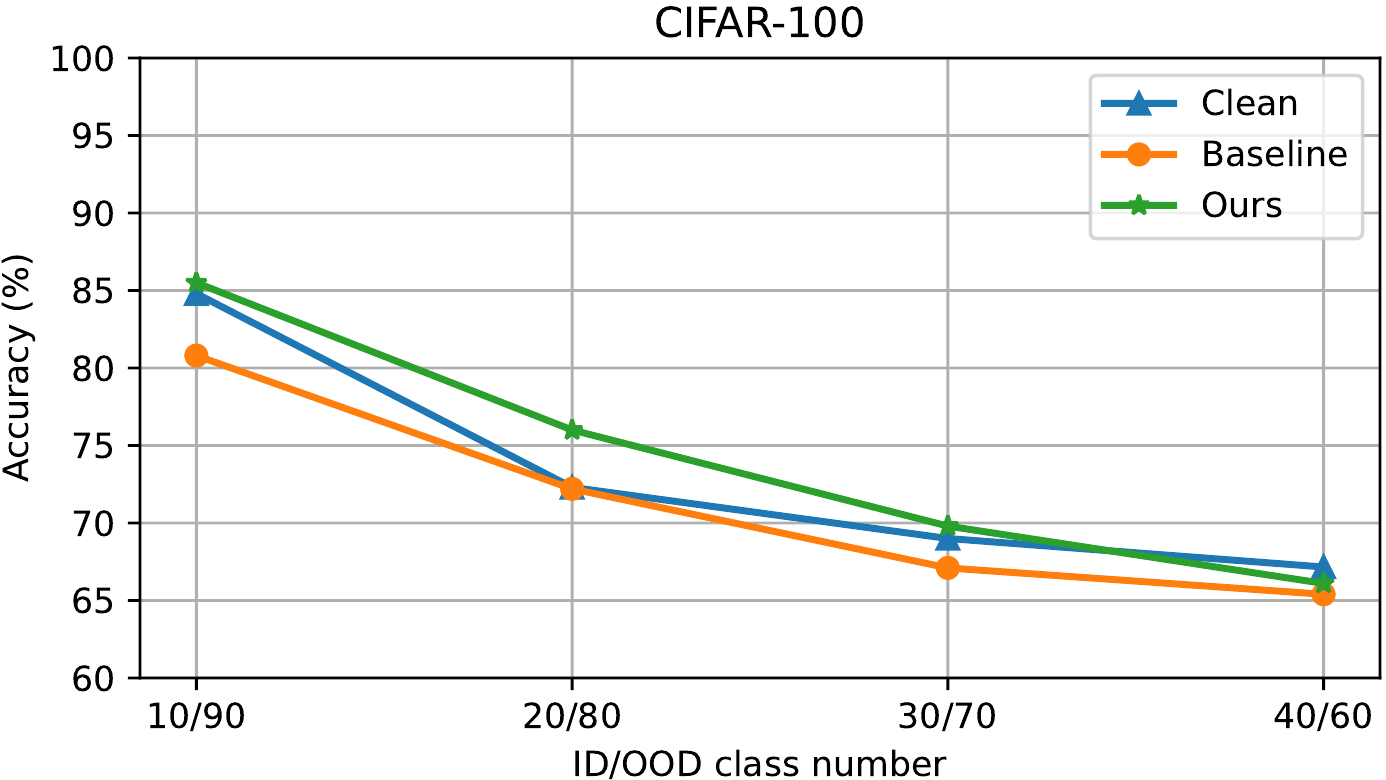}
    \end{center}
    
\caption{The performance of FixMatch~\cite{sohn2020fixmatch}, Clean and our method against four different ID/OOD ratios on CIFAR-100.}
\label{baseline_vs_oracle}
\end{figure}

\subsection{Robustness Against ID/OOD Ratios}

To demonstrate the robustness of our method against different ratios of ID/OOD samples in the training dataset, we test our method against the FixMatch~\cite{sohn2020fixmatch} baseline and its ``Clean'' variant on CIFAR-100~\cite{krizhevsky2009learning} dataset against four different ID/OOD ratios: 10/90, 20/80, 30/70 and 40/60. 
% In the paper we provide two kinds of ID/OOD ratio for each dataset. 
% In this section, we further investigate the performance of FixMatch\cite{sohn2020fixmatch}, Clean and our method with different ID/OOD ratios. 
% We adopt four kinds of settings, 10/90, 20/80, 30/70 and 40/60 of CIFAR-100\cite{krizhevsky2009learning} and report the average performance of three runs. 
We report the average performance over three runs. 
As Fig.~\ref{baseline_vs_oracle} shows, 
% the performance of ``Clean'' is not always higher than FixMatch baseline (e.g. 20/80). Compared to 10/90, 20/80 setting has more ID classes in classification task and thus decreases the performance of ``Clean'' model. On the other hand, the performance degradation of FixMatch is smaller since OOD data benefit the training of feature extractor when ID samples are not sufficient. 
our method outperforms the FixMatch baseline in all four settings and achieves higher accuracy than the Clean model in three settings: 10/90, 20/80 and 30/70. 
Such a constant improvement justifies the robustness of our method against different ID/OOD ratios. \textcolor{myblue}{Note that the degradation of performance in all three models is caused by the increasing ID class number.} 
% experiment demonstrates that our method can improve the performance of SSL model with different ID/OOD ratios. 
% Besides, it proves that OOD can improve SSL model if they are properly utilized in model training.

\subsection{Performance on ID/OOD Classification}
\textcolor{myblue}{
Following previous studies\cite{saito2021openmatch}, we also compare the performance of our method with those of previous open-set semi-supervised learning methods on ID/OOD classification. 
% we also compare the performance of our method on ID/OOD classification to previous open-set semi-supervised learning methods. 
The experiments are conducted on CIFAR-100 with two different settings and the AUROC values of each method are shown in Table~\ref{tab:auroc}. 
As shown in the table, despite the imbalance between ID and OOD samples, our method achieves a significant improvement over previous methods. 
Please note that we use the output probabilities of the predicted class as ID probabilities to compute the AUROC value of FixMatch.
% We also provide the AUROC value of FixMatch by directly using the output probabilities of the predicted class as its ID probability. 
}

\begin{table}
    \caption{The performance of our method against different numbers of pools (DomainNet-Real 20/50k).}
   \begin{center}
   \begin{tabular}{l c c c c c}
   \toprule
   No. of Pools & 0 & 1 & 2 & 3 & 4 \\
   \midrule
   Our Method & 52.5 & 53.2 & 54.3 & 52.2 & 50.8 \\
   \bottomrule
   \end{tabular}
   \end{center}

   \label{pool_num_table}
\end{table}

\begin{table}
    \caption{\textcolor{myblue}{The performance of our method against different $N_{id}$ (DomainNet-Real 20/50k).}}
   \begin{center}
   {\color{myblue}\begin{tabular}{l c c c c c}
   \toprule
   $N_{id}$ & 5 & 6 & 7 & 8 & 9 \\
   \midrule
   Our Method & 53.8 & 54.3 & 54.0 & 53.1 & 52.6 \\
   \bottomrule
   \end{tabular}}
   \end{center}

   \label{tab:N_id_ablation}
\end{table}
 
\subsection{Justification of our Choice on No. of Pools}

To justify our choice of using a cascade of two pools in importance-based sampling, we investigate how the number of pools influences the performance of our method (Table~\ref{pool_num_table}).
% In this part, we conduct experiments with different pool number on DomainNet-Real and investigate the effect. 
All other hyperparameters are kept the same across all experiments.
% Section~\ref{implementation_details}. 
% The result are shown in table~\ref{pool_num_table}.
It can be observed that: i) using two pools achieves the best performance;
ii) when using zero or one pool, the density of ID samples is not high enough and thus worsens the minibatch training;
iii) when using three or four pools, the density is improved but at the cost of filtering out too many unlabeled (ID) samples, which yields overfitting and also worsens the training.
Thus, we use a cascade of two pools in our importance-based sampling implementation.
% The model degenerates to clustering only when pool number is 0. It is obvious that reasonable pool number setting can improve the performance of SSL model by focusing on identified ID samples. But as pool number goes up, the density of ID samples reaches the peak and the SSL model  only concentrates on a small part of unlabeled data as most of them are filtered out by cascaded ID sample pools. We report our method's performance when pool number is 2.

\subsection{\textcolor{myblue}{Threshold of ID/OOD Identification}}
\textcolor{myblue}{
The ID/OOD identification of our method selects $N_{id}$ prototypes that are closest to the feature center of labeled samples as ID prototypes. To investigate the influence of $N_{id}$ selection, we test our method on DomainNet-Real 20/50k and the results are shown in Table~\ref{tab:N_id_ablation}. 
It can be observed that our method performs better than the baseline with different $N_{id}$ and the best performance is achieved at 6. 
}

\begin{table}
 \caption{The performance of our method against the number of prototypes (CIFAR-100 10/90).}
  \begin{center}
  \begin{tabular}{l c c c c c}
  \toprule
  Prototypes Num & 2 & 6 & 10 & 14 & 18 \\
  \midrule
  Our Method & 83.3 & 84.4 & 85.5 & 84.0 & 83.7 \\
  \bottomrule
  \end{tabular}
  \end{center}
   \label{prototypes_num_table}
\end{table}

\subsection{Number of Prototypes}
To investigate how the number of prototypes influences the performance of our method, we test different choices of it on the CIFAR-100 10/90 setting and show the results in Table~\ref{prototypes_num_table}.
It can be observed that our method is insensitive to the number of prototypes and almost always outperforms the baseline (FixMatch~\cite{sohn2020fixmatch}).
Thus, we suggest to set the default value of it as $10$ (as used in this paper).

\begin{figure}[!htb]
      \begin{center}
          \includegraphics[height=0.5\linewidth,width=0.7\linewidth]{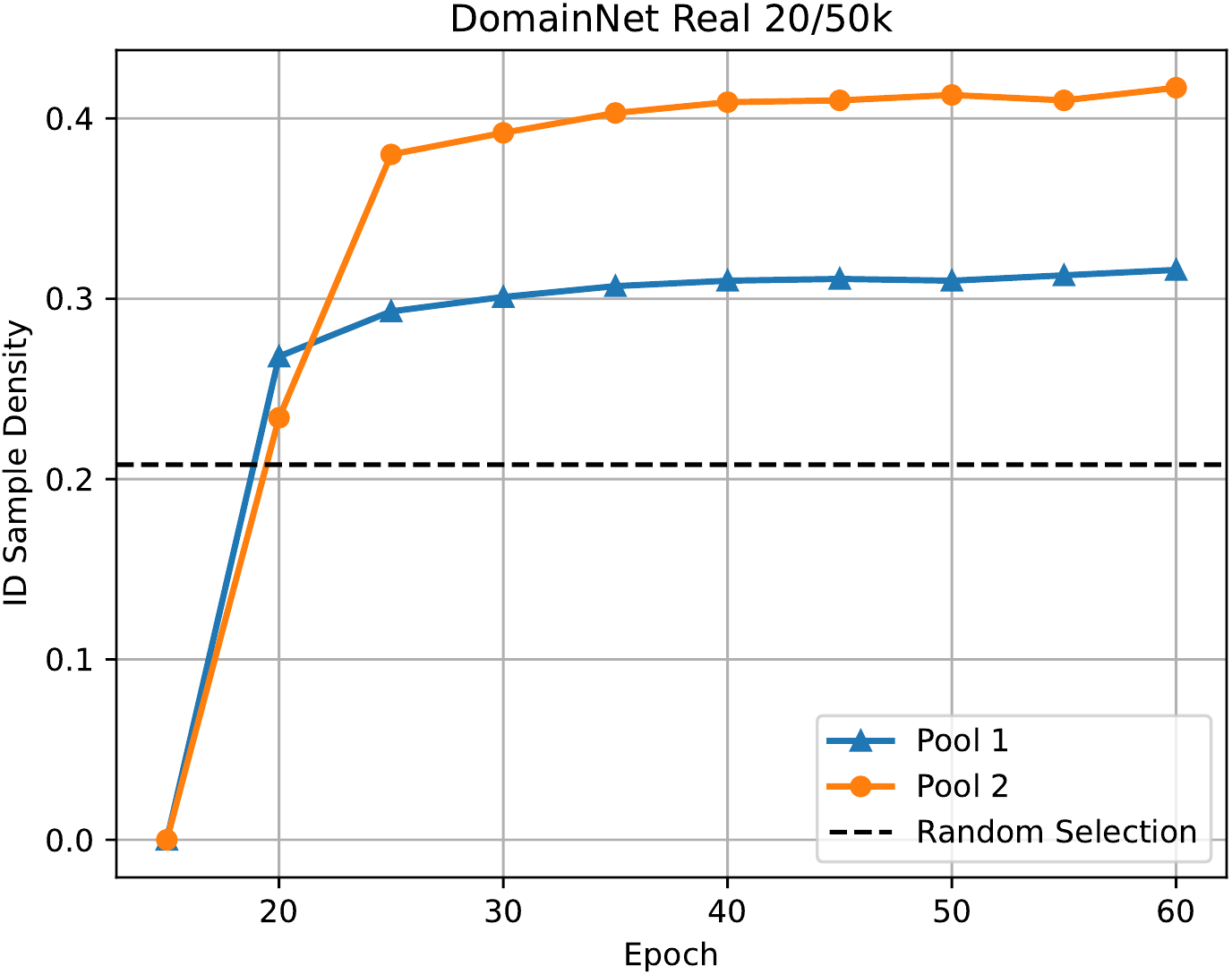}
      \end{center}
%   \caption{The improvement of ID density of two pools in DomainNet Real. The ID sample number is increasing as the model iterates. }
\caption{Justification of ID sample refinement (DomainNet -Real 20/50k). The ID sample density of our ID sample pools are much higher than that of the raw unlabeled dataset (Random Selection).}
   \label{id_density_figure}
\end{figure}

\subsection{Effectiveness of Cascading Pools}
To further justify the effectiveness of our cascading pooling strategy, we plot the density of ID samples in our two cascaded ID sample pools (with per-class capacity 300 and 150 respectively) against training epochs when training our model with the DomainNet-Real 20/50k setting (Figure~\ref{id_density_figure}).
We also marked the percentage of ID samples in the raw unlabeled dataset as ``Random Selection''. 
It can be observed that:
i) Our two pools have much higher ID sample densities than Random Selection (approximately 10\% and 20\% respectively), which justifies the usefulness of our approach.
ii) Pool 2 has a much higher ID sample density than Pool 1 (approximately 10\%), which indicates the effectiveness of our cascading pooling strategy.

% \subsection{Open-set Semi-Supervised Learning on Fine-grained Classification}
\subsection{\textcolor{myblue}{Performance on Fine-grained Classification}}

\textcolor{myblue}{
% Fine-grained classification\cite{} is a more challenging task than regular objection classification.
Fine-grained classification\cite{akata2015evaluation,yang2018learning,dubey2018maximum,syeda2020chest,zhu2019fine} aims to distinguish between objects that previously belong to the same (coarse-level) class, \eg, species of birds.
Recently, there have been some studies that apply open-set semi-supervised learning on fine-grained classification\cite{su2021realistic}, whose datasets contain both ID and OOD data. 
% Recently, there have been some studies that focus on open-set semi-supervised fine-grained learning\cite{}, where the fine-grained evaluation dataset contains both in-distribution and out-of-distribution data. 
This is a more challenging task as samples in fine-grained classes (\eg, different brands of cars) have less discriminative features.  
In this section, we verify the effectiveness of our method on fine-grained classification.
% The evaluation of fine-grained open-set semi-supervised learning is more difficult for ID/OOD classification, as both ID and OOD data come from the same domain and the difference between them is much smaller. 
}
% In this section, we verify the effectiveness of our method on the fine-grained open-set semi-supervised learning task.
% In this section, we verify the effectiveness of our method on fine-grained classification.

\vspace{2mm}
\noindent \textbf{\textcolor{myblue}{Datasets.}}
\textcolor{myblue}{
Following previous studies\cite{su2021realistic}, we evaluate our method on two fine-grained datasets that exhibit a long-tailed distribution of classes and contain a large number of out-of-class images: Semi-Aves (from the semi-supervised challenge at FGVC7 workshop\cite{su2021semi}) and Semi-Fungi (from the FGVC fungi challenge\cite{fungi_challenge}).
% We test the performance of our method on two datasets, Semi-Aves and Semi-Fungi. The two datasets, Semi-Aves and Semi-Fungi, are obtained by sampling classes from Aves(birds) taxonomy.
% Between them, Semi-Aves was part of the semi-supervised challenge at FGVC7 workshop\cite{}, while Semi-Fungi is from the FGVC fungi challenge\cite{}. 
% The first dataset was part of the semi-supervised challenge at FGVC7 workshop\cite{}, while the second dataset is sampled from the FGVC fungi challenge\cite{} following a similar scheme. 
The OOD images of both datasets are those that do not belong to the classes of the labeled set. 
% The OOD images of both datasets are the images not belonging to the classes of the labeled set. 
Between them, Semi-Aves contains 200 ID classes and 800 OOD classes, and 6K/27K/122K images in labeled set/ID unlabeled set/OOD unlabeled set, respectively. Semi-Fungi contains 200 ID classes and 1194 OOD classes, and 4K/13K/65K images in labeled set/ID unlabeled set/OOD unlabeled set, respectively.
Following \cite{su2021realistic}, we use the labeled and unlabeled set (containing both ID and OOD samples) provided by these datasets and ResNet-50\cite{he2016deep} as the backbone network for evaluation.
All samples are resized to a resolution of 224$\times$224 in all experiments.
% We used the labeled and unlabeled set provided by the datasets and use ResNet-50\cite{} as the backbone network for evaluation following \cite{}. 
% The unlabeled set contains both ID and OOD samples defined by the datasets, and all samples are resized to 224$\times$224 for all experiments. The experiment results are shown in Table~\ref{} and Table~\ref{}.
}

\vspace{2mm}
\noindent \textbf{\textcolor{myblue}{Comparison Setup.}}
\textcolor{myblue}{
Following\cite{su2021realistic}, we compare our method to the following competitors: 
% i) Supervised baseline, which trains the model with the labeled set only. 
i) Supervised baseline, where the model is trained only on the labeled set;
% ii) Pseudo-Labeling\cite{}, which uses a base model's confident prediction on unlabeled images as labels, and then trains a new model by sampling half of the batch from labeled data and half from unlabeled data during training. 
ii) Pseudo-Labeling\cite{lee2013pseudo}, which uses a base model's confident prediction on unlabeled images as pseudo-labels, and then trains a new model by sampling half of the batch from labeled data and half from pseudo-labeled data;
iii) Curriculum Pseudo-Labeling\cite{cascante2021curriculum}, which repeats the following for 5 times: training a supervised model with labeled data, and expanding labeled data by including ($\{20, 40, 60, 80, 100\}\%$ of) the unlabeled data with the highest predictions. 
% iii) Curriculum Pseudo-Labeling\cite{}. Curriculum Pseudo-Labeling generates the pseudo labels for unlabeled samples after the training on the labeled set is finished. First it trains the supervised baseline with labeled data, then add the unlabeled data with the highest predictions as \textit{new} labeled data to the labeled set. The training/adding procedure is repeated 5 times and labels $\{20, 40, 60, 80, 100\}\%$ of  unlabeled data respectively. 
iv) FixMatch\cite{sohn2020fixmatch};
v) Self-Training, which first trains a teacher model with the labeled set, and then trains a student model with a scaled cross-entropy loss on the unlabeled data and a cross-entropy loss on the labeled data. 
% v) Self-Training\cite{}. It first trains a teacher model with the labeled set, then trains a student model with scaled cross-entropy loss on the unlabeled data and cross-entropy loss on labeled data. 
vi) Supervised Oracle, which trains the model with the labeled set and ID unlabeled set with ground-truth labels.
}

\textcolor{myblue}{
% As shown in Table~\ref{} and Table~\ref{}, our method significantly outperforms previous methods and FixMatch. 
As shown in Table~\ref{tab:semi-aves} and Table~\ref{tab:semi-fungi}, our method significantly outperforms all previous methods, which demonstrates the effectiveness of our method on fine-grained classification.
% Although FixMatch is not the method with the best performance in previous studies, our method still achieves significant improvement in both top-1 and top-5 accuracy. 
% The superior performance demonstrates the effectiveness of our method in more challenging scenarios. 
}

\begin{table}
\caption{\textcolor{myblue}{Results on Semi-Aves benchmark. We experiment with six different SSL methods as well as supervised baselines. Results of other methods are copied from \cite{su2021realistic}.}}
   \begin{center}
   {\color{myblue}\begin{tabular}{l c c}
   \toprule
   Method & Top-1 & Top-5 \\
   \midrule
   Supervised baseline & 20.6±0.4 & 41.7±0.7 \\
   \midrule
   Pseudo-Label\cite{lee2013pseudo} & 12.2±0.8 & 31.9±1.6 \\
   Curriculum Pseudo-Label\cite{cascante2021curriculum} & 20.2±0.5 & 41.0±0.9 \\
   FixMatch\cite{sohn2020fixmatch} & 19.2±0.2 & 42.6±0.6 \\
   Self-Training & 22.0±0.5 & 43.3±0.2 \\
   \midrule
   Ours & 26.9±0.5 & 48.4±0.8 \\
   \midrule
   Supervised oracle & 57.4±0.3 & 79.2±0.1 \\
   \bottomrule
   \end{tabular}}
   \end{center}
   \label{tab:semi-aves}
\end{table}

\begin{table}
\caption{\textcolor{myblue}{Results on Semi-Fungi benchmark. We experiment with six different SSL methods as well as supervised baselines. Results of other methods are copied from \cite{su2021realistic}.}}
   \begin{center}
   {\color{myblue}\begin{tabular}{l c c}
   \toprule
   Method & Top-1 & Top-5 \\
   \midrule
   Supervised baseline & 31.0±0.4 & 54.7±0.8 \\
   \midrule
   Pseudo-Label\cite{lee2013pseudo} & 15.2±1.0 & 40.6±1.2 \\
   Curriculum Pseudo-Label\cite{cascante2021curriculum} & 30.8±0.1 & 54.4±0.3 \\
   FixMatch\cite{sohn2020fixmatch} & 25.2±0.3 & 50.2±0.8 \\
   Self-Training & 32.5±0.5 & 56.3±0.3 \\
   \midrule
   Ours & 34.4±0.4 & 58.0±0.8 \\
   \midrule
   Supervised oracle & 60.2±0.8 & 83.3±0.9 \\
   \bottomrule
   \end{tabular}}
   \end{center}
   \label{tab:semi-fungi}
\end{table}

\subsection{Visualization of ID/OOD Features}

\begin{figure}
\centering
\subfloat[Baseline]{{\includegraphics[width=.3\textwidth]{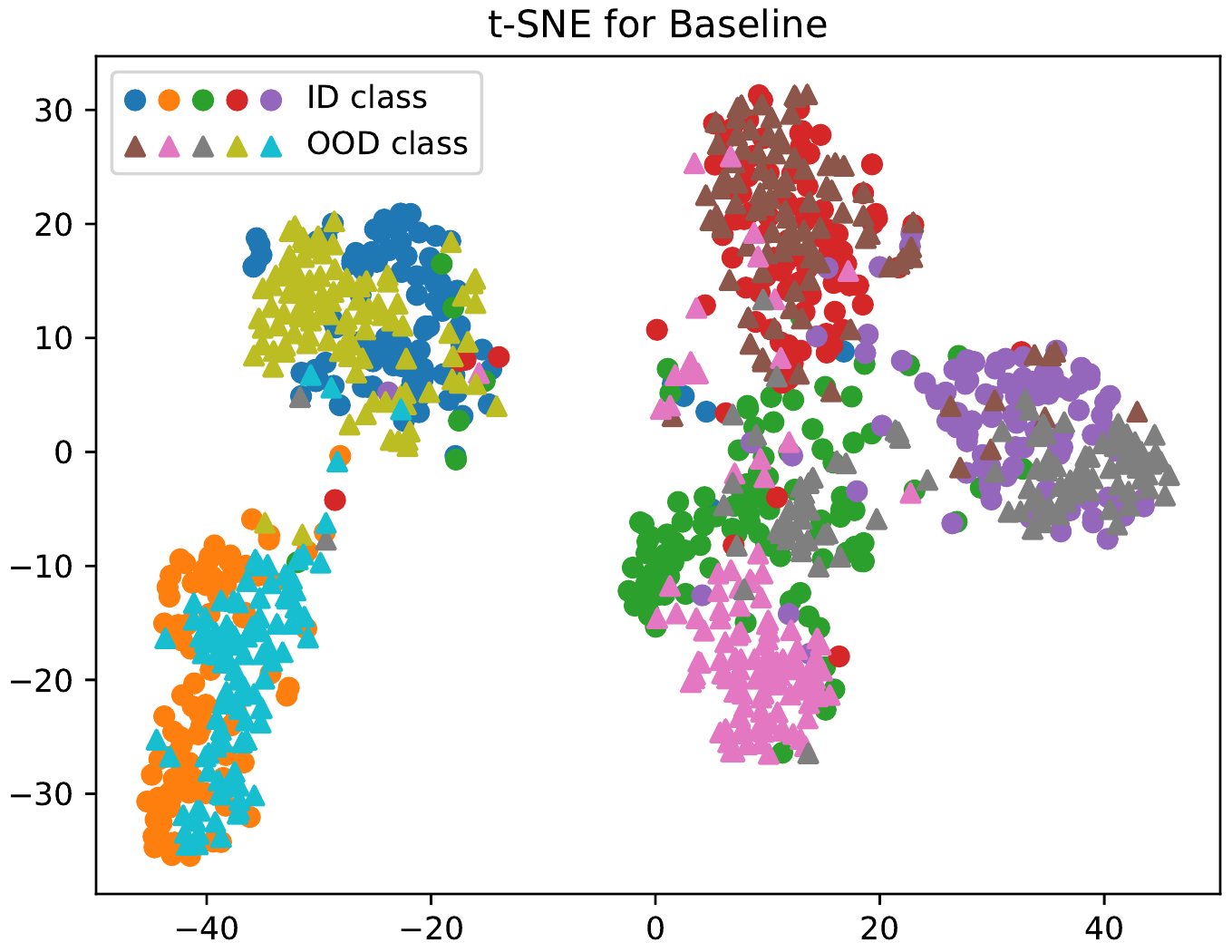} }\label{fig:baseline_tsne}}

\subfloat[Ours]{{\includegraphics[width=.3\textwidth]{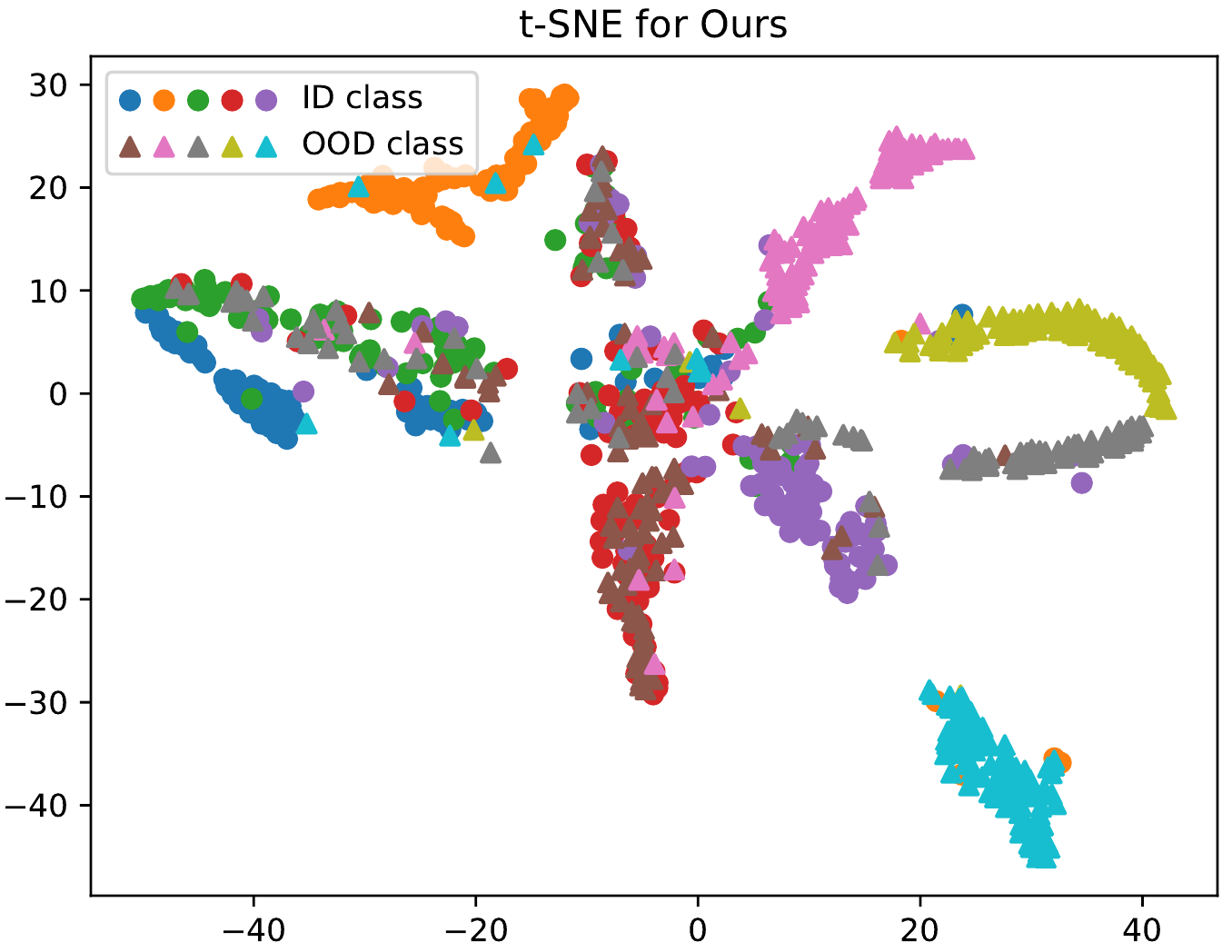} }\label{fig:ours_tsne}}
\caption{The t-SNE feature visualization on CIFAR-10. (a) provides the feature visualization of FixMatch. (b) shows the t-SNE results of our method. As shown in the figures, our method organizes both ID and OOD features and improves the feature extractor.}
\label{fig:tsne}
\end{figure}

In this section, we apply our method to CIFAR10 for feature visualization. To better illustrate the difference and distribution of ID/OOD features, we select all 10 classes in CIFAR10\cite{krizhevsky2009learning} rather than datasets with more categories in our experiment. We set the first 5 classes in CIFAR10 as ID and other 5 classes as OOD. The feature visualization is shown in Fig~\ref{fig:tsne}, including the visualization of both baseline (FixMatch) and our method. As shown in Fig~\ref{fig:baseline_tsne}, baseline model can not separate the ID and OOD features and thus confuses the OOD detector. However, in Fig~\ref{fig:ours_tsne}, our method can better cluster both ID/OOD features and thus preserves the difference between ID and OOD features in the feature level. Therefore, our method facilitates the training of the feature extractor and the ID/OOD classification in the importance-based sampling.

\section{Conclusion}

% In this paper, we find that the proper use of OOD samples can benefit semi-supervised learning (SSL) and propose two interdependent techniques for open-set SSL, including a prototype-based clustering algorithm and a sample refinement strategy.
In this paper, we find that the proper use of OOD samples can benefit semi-supervised learning (SSL).
% Accordingly, we propose two interdependent techniques for open-set SSL, including a prototype-based clustering algorithm and a sample refinement strategy.
Accordingly, we propose two techniques for open-set SSL: i) a prototype-based clustering and identification algorithm and ii) an importance-based sampling method.
% Prototype-based clustering algorithm clusters samples at feature-level and thus achieves better identification of ID and OOD samples by increasing their  distance.
% Our prototype-based clustering algorithm clusters samples at feature-level and thus achieves better identification of ID and OOD samples by increasing their distances in-between.
Our prototype-based clustering and identification algorithm clusters samples at feature-level and thus achieves better identification of ID and OOD samples by increasing their distances in-between.
% Our prototype-based clustering algorithm clusters samples at feature-level and thus achieves better identification of ID and OOD samples by increasing their distance.
Addressing the sampling bias introduced by the ID/OOD identification process, we propose an importance-based sampling method that maintains a pyramid of sample pools containing samples that are important to SSL.
% Considering that the low density of ID samples in a mini-batch is the culprit for significant model degradation, we propose a sample refinement strategy which maintains a pyramid of sample pools for ID sample distillation.
% Considering that the low density of ID samples in a mini-batch is the culprit leading to significant model degradation, a sample refinement strategy which maintains a pyramid of sample pools is designed for ID samples purification.
We implemented our method on top of FixMatch~\cite{sohn2020fixmatch} and achieved state-of-the-art in open-set SSL on extensive public benchmarks.
% Our proposed method is implemented on top of FixMatch~\cite{sohn2020fixmatch} and achieves state-of-the-art in open-set SSL on extensive public benchmarks.

%In this paper we propose a new method for open-set semi-supervised learning. Out-Of-Distribution (OOD) data can benefit the semi-supervised learning model if they take part in the training procedure properly or ID data is not sufficient. We involve the OOD data in training with prototypes-based clustering and leverage both ID and OOD unlabeled data to obtain a better feature extractor. Later, we identified ID and OOD samples based on the prototype-based clustering result and refine the unlabeled dataset to improve the density of ID samples. Our experiment demonstrate that our method can improve the performance of SSL model in several representative open-set SSL setting.

\iffalse
% use section* for acknowledgment
\ifCLASSOPTIONcompsoc
  % The Computer Society usually uses the plural form
  \section*{Acknowledgments}
\else
  % regular IEEE prefers the singular form
  \section*{Acknowledgment}
\fi
The authors would like to thank...
\fi

% Can use something like this to put references on a page
% by themselves when using endfloat and the captionsoff option.
\ifCLASSOPTIONcaptionsoff
  \newpage
\fi

% trigger a \newpage just before the given reference
% number - used to balance the columns on the last page
% adjust value as needed - may need to be readjusted if
% the document is modified later
%\IEEEtriggeratref{8}
% The "triggered" command can be changed if desired:
%\IEEEtriggercmd{\enlargethispage{-5in}}

% references section

% can use a bibliography generated by BibTeX as a .bbl file
% BibTeX documentation can be easily obtained at:
% http://mirror.ctan.org/biblio/bibtex/contrib/doc/
% The IEEEtran BibTeX style support page is at:
% http://www.michaelshell.org/tex/ieeetran/bibtex/
\bibliographystyle{IEEEtran}
% argument is your BibTeX string definitions and bibliography database(s)
\bibliography{egbib}

\vspace{-10mm}
\begin{IEEEbiography}[{\includegraphics[width=1in,height=1.25in,clip,keepaspectratio]{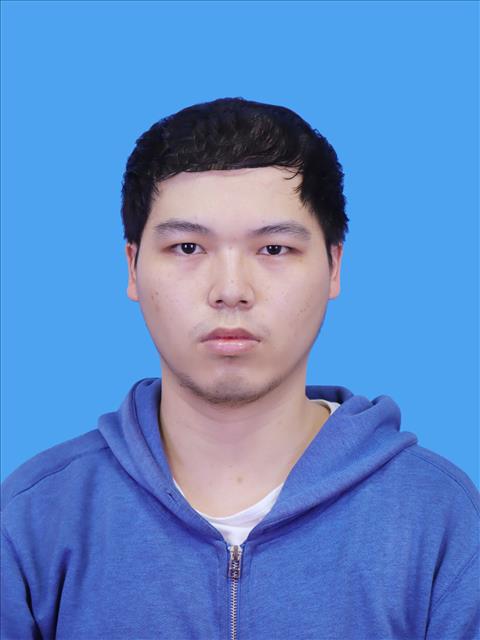}}]{Ganlong Zhao} received the BEng and MS degrees from the School of Computer Science and Engineering of Sun Yat-Sen University in 2019 and 2021 respectively. He is currently working toward the PhD degree in the Department of Computer Science at The University of Hong Kong. He worked as a research intern at Meituan Inc., in 2020. His research interests include machine learning and computer vision, specifically semi-supervised and unsupervised learning. 
\end{IEEEbiography}

\vspace{-10mm}
\begin{IEEEbiography}[{\includegraphics[width=1in,height=1.25in,clip,keepaspectratio]{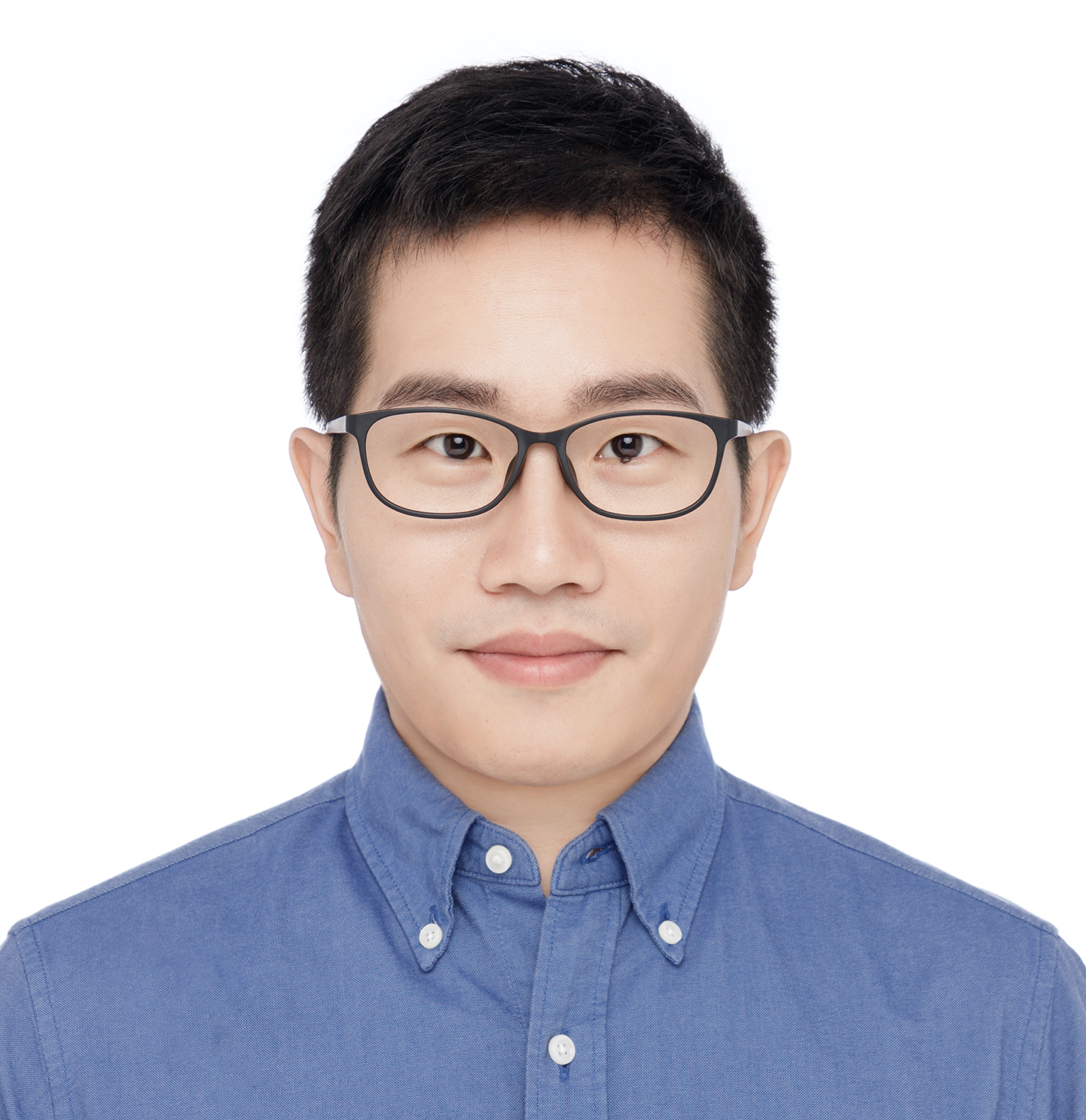}}]{Guanbin Li} (M'15) is currently an associate professor in School of Data and Computer Science, Sun Yat-sen University. He received his PhD degree from the University of Hong Kong in 2016. His current research interests include computer vision, image processing, and deep learning. He is a recipient of ICCV 2019 Best Paper Nomination Award. He has authorized and co-authorized on more than 100 papers in top-tier academic journals and conferences. He serves as an area chair for the conference of VISAPP. He has been serving as a reviewer for numerous academic journals and conferences such as TPAMI, IJCV, TIP, TMM, TCyb, CVPR, ICCV, ECCV and NeurIPS.
\end{IEEEbiography}

\vspace{-10mm}
\begin{IEEEbiography}[{\includegraphics[width=1in,height=1.25in,clip,keepaspectratio]{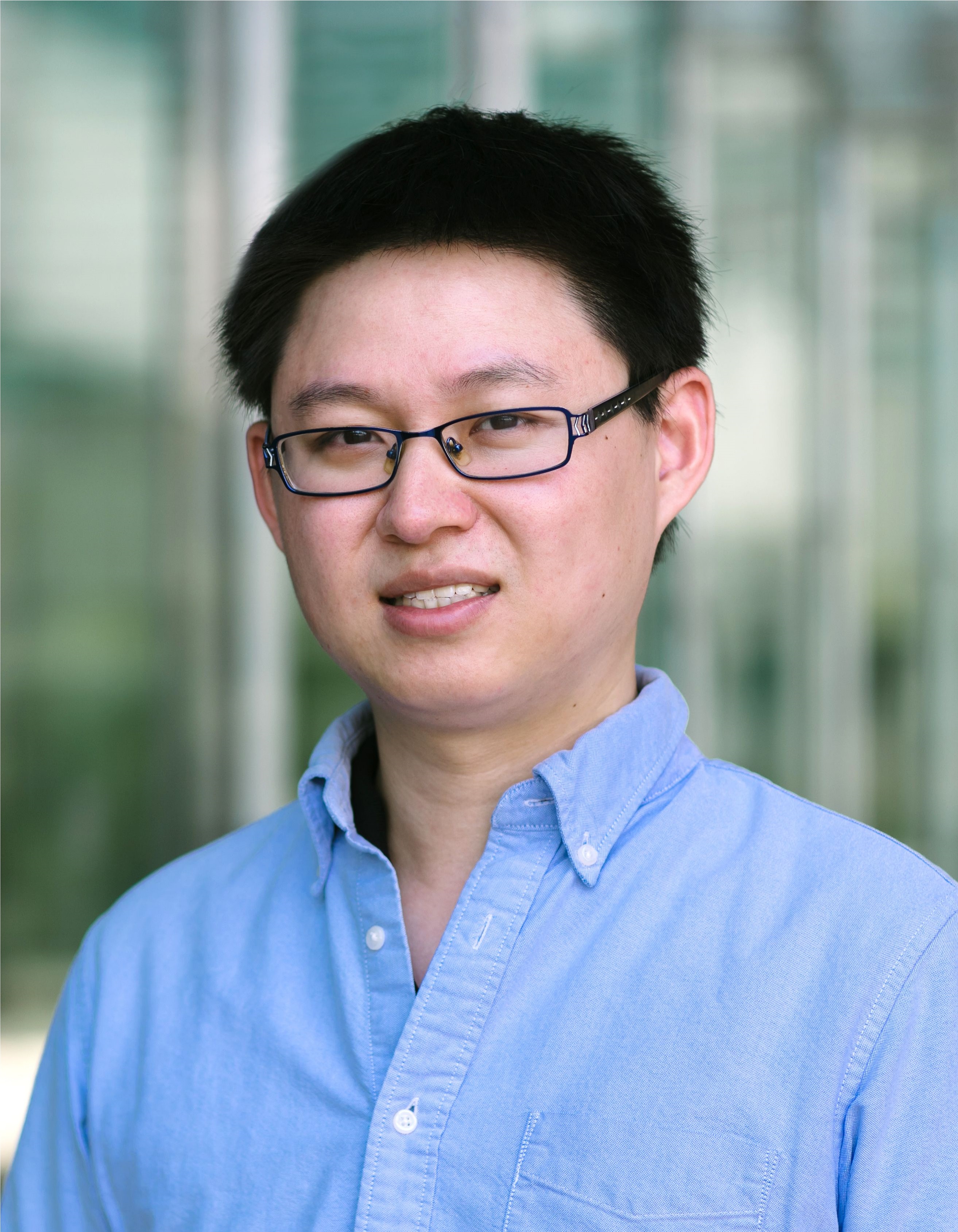}}]{Yipeng Qin} received the BS degree from Shanghai Jiaotong University in 2013, and the PhD degree from National Centre for Computer Animation (NCCA), Bournemouth University in 2017. From 2017 to 2019 he was a Postdoctoral Research Fellow at the Visual Computing Center (VCC), King Abdullah University of Science and Technology (KAUST). He joined the School of Computer Science and Informatics, Cardiff University, in 2019 as a lecturer. His current research interests include machine learning, computer vision and computer graphics.
\end{IEEEbiography}

\vspace{-10mm}
\begin{IEEEbiography}[{\includegraphics[width=1in,height=1.25in,clip,keepaspectratio]{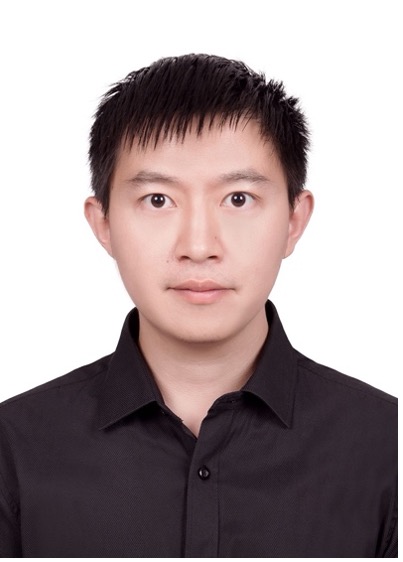}}]{Jinjin Zhang} obtained M.Eng. in computer science from Beihang University in 2017 and B.E. degree from Changchun University of Science and Technology in 2013. He is now a senior computer vision engineer in Meituan. His research interests include self-supervised learning, semi-supervised learning and its applications to computer vision.
\end{IEEEbiography}

\vspace{-10mm}
\begin{IEEEbiography}[{\includegraphics[width=1in,height=1.25in,clip,keepaspectratio]{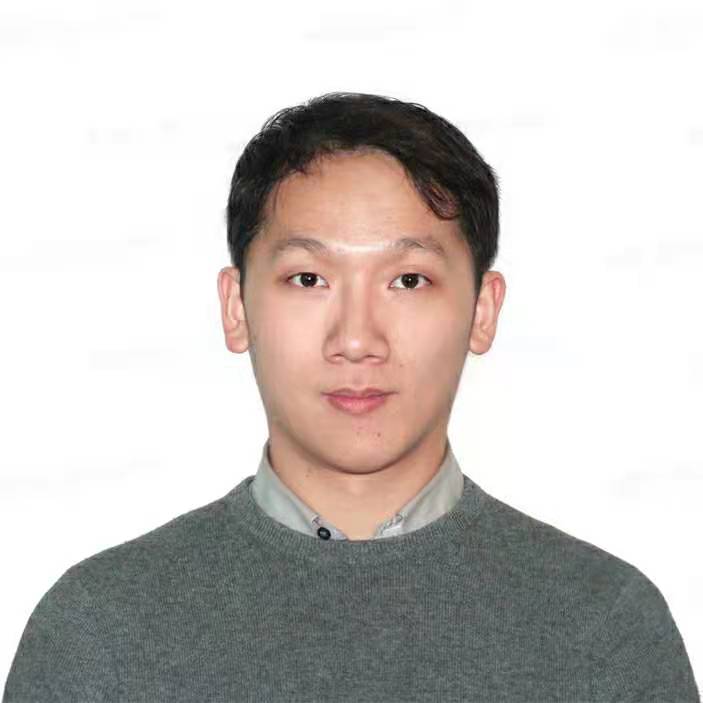}}]{Zhenhua Chai} received the B.E. (Hons.) degree in automation from the Central University of Nationality and the Ph.D. degree in computer application technology from the National Lab of Pattern Recognition, Institute of Automation, Chinese Academy of Sciences in 2008 and 2013, respectively. His research interests focus on AutoDL, model compression, self supervised learning and applications on face analysis. Currently he works as a research expert in Meituan.
\end{IEEEbiography}

\vspace{-10mm}
\begin{IEEEbiography}[{\includegraphics[width=1in,height=1.25in,clip,keepaspectratio]{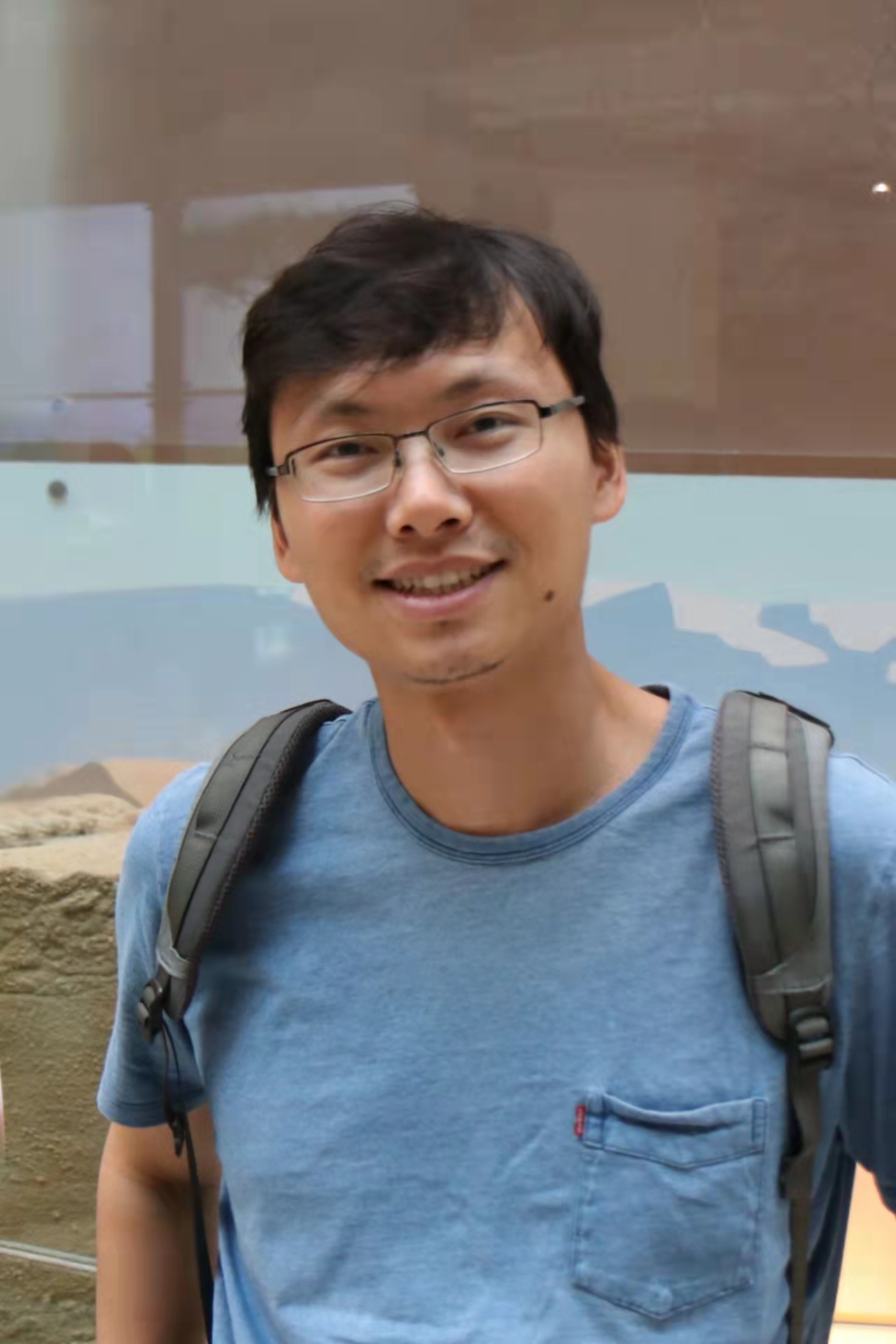}}]{Xiaolin Wei} received Ph.D. in Computer Science from Texas A\&M University. His research area includes computer vision, machine learning, computer graphics, 3D vision, augmented reality. He worked as a research engineer at Google, Virtroid and Magic Leap, and now is working at Meituan AI Lab.
\end{IEEEbiography}

\begin{IEEEbiography}[{\includegraphics[width=1in,height=1.25in,clip,keepaspectratio]{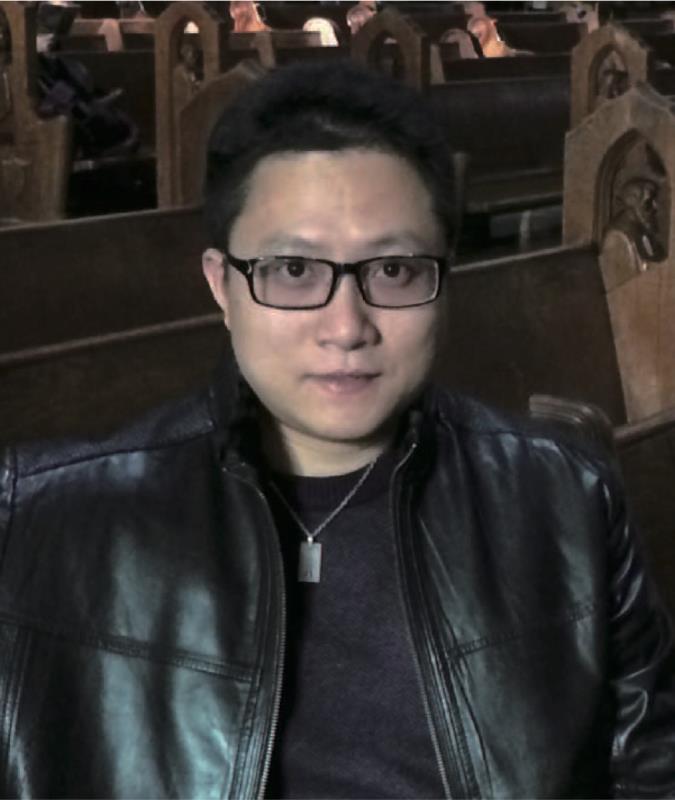}}]{Liang Lin} (M'09, SM'15) is a full Professor of Sun Yat-sen University. He is an Excellent Young Scientist of the National Natural Science Foundation of China. From 2008 to 2010, he was a Post-Doctoral Fellow at the University of California, Los Angeles. From 2014 to 2015, as a senior visiting scholar, he was with The Hong Kong Polytechnic University and The Chinese University of Hong Kong. He currently leads the SenseTime R$\&$D teams to develop cutting-edge and deliverable solutions on computer vision, data analysis and mining, and intelligent robotic systems. He has authored and co-authored more than 100 papers in top-tier academic journals and conferences. He has been serving as an associate editor of IEEE Trans. Human-Machine Systems, The Visual Computer and Neurocomputing. He served as area/session chairs for numerous conferences, such as ICME, ACCV, ICMR. He was the recipient of the Best Paper Runners-Up Award in ACM NPAR 2010, the Google Faculty Award in 2012, the Best Paper Diamond Award in IEEE ICME 2017, and the Hong Kong Scholars Award in 2014. He is a Fellow of IET.
\end{IEEEbiography}

\vspace{-10mm}
\begin{IEEEbiography}[{\includegraphics[width=1in,height=1.25in,clip,keepaspectratio]{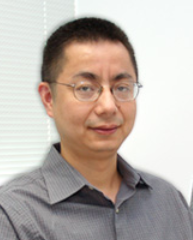}}]{Yizhou Yu} received the PhD degree from University of California at Berkeley in 2000. He is currently a professor at The University of Hong Kong, and was a faculty member at University of Illinois, Urbana-Champaign between 2000 and 2012. He is a recipient of 2002 US National Science Foundation CAREER Award, and 2007 NNSF China Overseas Distinguished Young Investigator Award. He has served on the editorial board of IET Computer Vision, IEEE Transactions on Visualization and Computer Graphics, The Visual Computer, and International Journal of Software and Informatics. He has also served on the program committee of many leading international conferences, including SIGGRAPH, SIGGRAPH Asia, and International Conference on Computer Vision. His current research interests include deep learning methods for computer vision, computational visual media, geometric computing, video analytics and biomedical data analysis.
\end{IEEEbiography}

% that's all folks
\end{document}